\pdfoutput=1
\documentclass[]{article}

\usepackage{longtable}
\usepackage{booktabs}
\usepackage{todonotes} 
\usepackage{caption}
\usepackage{subfig}
\usepackage{graphicx}
\usepackage{microtype}
\usepackage{amsmath}
\usepackage{authblk}
\usepackage{natbib}
\usepackage{arxiv}
\title{Uncertainty estimation for classification and risk prediction on medical tabular data}

\author{Lotta Meijerink}
\author{Giovanni Cin\`a}
\author{Michele Tonutti}
\affil{Pacmed, Amsterdam}

\begin{document}
\maketitle
\begin{abstract}
In a data-scarce field such as healthcare, where models often deliver predictions on patients with rare conditions, the ability to measure the uncertainty of a model's prediction could potentially lead to improved effectiveness of decision support tools and increased user trust. 

This work advances the understanding of uncertainty estimation for classification and risk prediction on medical tabular data, in a two-fold way. First, we expand and refine the set of heuristics to select an uncertainty estimation technique, introducing tests for clinically-relevant scenarios such as generalization to uncommon pathologies, changes in clinical protocol and simulations of corrupted data. We furthermore differentiate these heuristics depending on the clinical use-case. Second, we observe that ensembles and related techniques perform poorly when it comes to detecting out-of-domain examples, a critical task which is carried out more successfully by auto-encoders.
These remarks are enriched by considerations of the interplay of uncertainty estimation with class imbalance, post-modeling calibration and other modeling procedures. Our findings are supported by an array of experiments on toy and real-world data.
\end{abstract}

\section{Introduction}

Automated diagnosis via classification is the first FDA-approved medical application of Machine Learning  \citep{keane2018eye} and is poised to transform healthcare in the coming decades. At the same time, risk prediction models are among the most common prognostic models in healthcare research: a staggering amount of risk models has been developed in different subfields of medicine, to predict the probability of adverse events ranging from in-hospital mortality to post-surgical incontinence.\footnote{As an example, \cite{damen2016prediction} report that until 2013 at least 363 prediction models had been developed for the specific task of predicting the cardiovascular disease risk in the general population.} The reason for this proliferation is that an accurate risk assessment can aid decision making and potentially lead to great improvements in patient outcomes \citep{jaspers2011effects}. To date, risk prediction models are incorporated in numerous clinical protocols and form the basis of several guidelines (see e.g. \cite{mosca2011effectiveness}).

There is however a potential issue that is often not addressed in the development of said models, namely that the risk score, already reflecting uncertainty about the outcome, does not necessarily capture the uncertainty due to lack of data. This problem is particularly felt in medical applications, where the size of data sets is often limited: models with good overall performance are nonetheless likely to give unreliable prediction on uncommon patients. Indicating the level of confidence in a prediction, or outright discarding the predictions that are too uncertain, could vastly improve  the effectiveness of decision support tools as well as increase the trust of users in models' suggestions.

A growing body of literature in the field of Machine Learning is concerned with the development of techniques to estimate the uncertainty of individual predictions. These efforts were mostly directed towards regression tasks, with less attention devoted to classification and risk prediction. Moreover, among the few medical applications, most of them focused on image data. To enable the use of uncertainty estimation techniques for a wider range of prognostic and diagnostic models, what is required is a systematic array of experiments demonstrating which techniques work in practice for classification and risk prediction on tabular data, the format in which the vast majority of medical data is stored. The goal of this paper is to fill this gap and provide practical insights that can be used in the development of medical AI.

Intuitively, there are (at least) two high-level desiderata for uncertainty estimation techniques. First, the correlation between uncertainty and performance, in the sense that a model should perform better on groups with lower uncertainty.\footnote{This is sometimes called the risk-coverage trade-off \citep{geifman2017selective}.} Second, the effective detection of unforeseen and strange data points, meaning that uncertainty should increase on so-called out-of-distribution examples.
Employing these criteria, we evaluate the benefits and limitations of two different but intuitive frameworks to incorporate uncertainty in a risk prediction model in healthcare, running several experiments on toy and real-world tabular data. For medical data, we focus on the task of mortality prediction in the Intensive Care (IC). The code employed in said experiments is available online and results are fully reproducible. In what follows we summarize our contributions.

\subsection*{Generalizable Insights about Machine Learning in the Context of Healthcare}
\begin{enumerate}
\item we expand and refine existing heuristics to decide if and how to employ an uncertainty estimation technique, distinguishing between different clinical use-cases; for detection of out-of-domain examples we address multiple clinically-relevant scenarios such as generalization to different ethnic subgroups and uncommon pathologies.
\item we analyze the interaction of uncertainty estimation with other common modeling procedures such as class weighting and post-modeling calibration;
\item even if ensembling still is the top performer when it comes to the correlation between uncertainty and performance on test data,  we observe that, in contrast to previous research, it performs poorly at flagging out-of-domain examples, a task for which novelty detection methods seem to be better equipped.
\end{enumerate}

\section{Related Work}
Uncertainty is often divided into two types: aleatory uncertainty and epistemic uncertainty. The former refers to the intrinsic randomness of a phenomenon, whereas the latter is presumed to be caused by lack of knowledge \citep{der2009aleatory}.

\begin{figure}[!ht]
    \centering
    \includegraphics[width=0.4\linewidth]{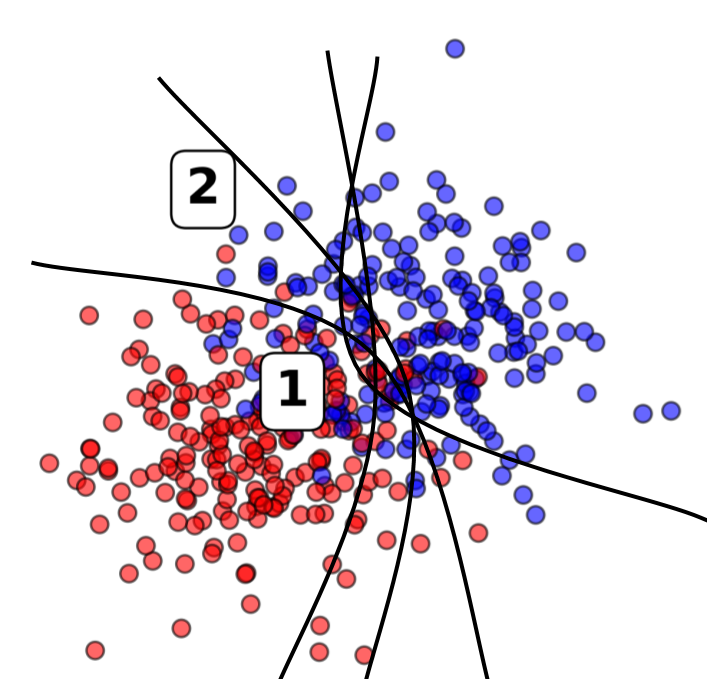}
    \caption{Illustration of the difference between aleatory and epistemic uncertainty in a classification problem. In the area indicated by (1) there is inherent overlap (aleatory uncertainty). In (2), where there are many different plausible decision boundaries, there is more uncertainty due to lack of knowledge about the optimal predictor (epistemic uncertainty).}
    \label{fig:aleatoryvsepistemic}
\end{figure}

\paragraph{Aleatory uncertainty}
In Machine Learning (ML henceforth) literature, aleatory uncertainty is interpreted as uncertainty due to ambiguity or noise in the data. In a classification problem this translates to having inherently overlapping classes. Therefore, such uncertainty cannot be resolved by observing more data points from the same source, but only by adding extra features, improving the quality of existing features, or other operations that could possibly make the classes separable. In medical prognostic models we almost always face a certain degree of aleatory uncertainty, as we do not have perfect information to predict future events. 
\paragraph{Epistemic uncertainty} On the other hand, epistemic uncertainty can be interpreted as uncertainty due to a lack of knowledge about the optimal predictor. This can be divided into uncertainty over model parameters and uncertainty over model structure (or hypothesis class).\footnote{In ML literature it often happens that only uncertainty over model parameters is considered and referred to as epistemic uncertainty.} Epistemic uncertainty can be resolved by observing more data, and is therefore also known as reducible uncertainty \citep{urbina2011quantification}. 
In Figure \ref{fig:aleatoryvsepistemic} we illustrate the difference between uncertainty due to inherent overlap of the classes, and uncertainty due to lack of knowledge about the optimal predictor. 

\paragraph{Modeling uncertainty}
There has been much work on modeling aleatory uncertainty in a classification problem, as this is what standard probabilistic classifiers are targeting by modeling the probabilities $p(y|x)$.  However, these probabilities often result in unjustified high confidence for points far away from the training data and thus fail to represent epistemic uncertainty \citep{gal2016dropout}.  
An intuitive way to better capture epistemic uncertainty is by modeling uncertainty over possible models or model parameters, which we will refer to as \textit{parameter uncertainty}. This builds on the idea that plausible models, which we can conceptualize as different `experts', will have different opinions about points far away from the training data. One can model uncertainty over model parameters by learning a posterior distribution over said parameters by (approximate) Bayesian modeling \citep{gal2016dropout, uncertaintytutorial}.  Applications in healthcare, e.g. \cite{benchmarkingbayesian, leibig2017leveraging}, showed that we can benefit from Bayesian deep learning techniques to incorporate epistemic uncertainty for detecting diabetic retinopathy from fundus photos, while \cite{ruhe2019bayesian} examined their application to mortality prediction in Intensive Care. However, other approaches such as linearly combining predictions into ensembles, have also shown to work well at capturing epistemic uncertainty \citep{fumera2004analysis, osband2016deep} and are able to outperform Bayesian methods \citep{lakshminarayanan2017simple, snoek2019can}.

Another straightforward way to estimate epistemic uncertainty is by simply using a separate model to give an indication of novelty or rarity, for example unconditional density estimation \citep{bishop1994novelty, markou2003novelty}.  We will refer to this type of uncertainty as \textit{novelty detection}. This paradigm relies on the idea that a model's predictions are likely to be less reliable in low density areas, since the model has no `experience' with similar data.
Additionally, other lines of research model aleatory and epistemic uncertainty with fuzzy preference modeling \citep{senge2014reliable}, possibility theory and evidence theory \citep{uncertaintytutorial}. Other methods involve retraining a model with new data and comparing the difference between the trained and retrained model \citep{kukar2002reliable}. Furthermore, there are approaches that train models end-to-end while optimizing for the exclusion of a portion of uncertain data \citep{geifman2019selectivenet}.
Finally, there are methods  leveraging the disagreement of a classifier with another model such as a generative model or a nearest neighbour classifier \citep{myers2020identifying, jiang2018trust}.

A recent study \citep{snoek2019can} tested a wide array of methods on tabular data, suggesting that ensembles are the best techniques for detecting covariate shift, namely a form of out-of-domain where features are shifted with respect to training data. The tests on tabular data, however, only comprised one dataset, and the covariate shift was emulated by randomly flipping categorical variables. Albeit informative, this scenario that does not encompass all the subtle ways in which medical data can change: due to the interactions between bodily functions and change in care protocol, it is likely that new patient groups will exhibit novel combination of features. It is therefore required to test said techniques on medical data more extensively, and to vastly expand the out-of-domain tests with clinically-informed experiments.

\section{Methods}
We decided to focus on two techniques that are most intuitive (let us not forget that uncertainty must eventually be interpretable for clinicians), allow us to work with any powerful discriminative model, do not require using training data during prediction, and require little effort to incorporate in a modeling process.
We investigate ensembles of probabilistic classifiers as representatives of \textit{parameter uncertainty}. These methods have been shown to be at least on par with Bayesian techniques in terms of performance \citep{filos2019systematic, lakshminarayanan2017simple, snoek2019can}, and are intuitive and straightforward to implement. Furthermore, there is the advantage that ensembling can be used in combination with any probabilistic classifier. We implement a variational autoencoder as a representative of a \textit{novelty detection} method \citep{vae}. To put our results in context, we also report results for a number of known benchmarks. 

\subsection{Models}

\subsubsection{Deep ensemble}
We implemented an ensemble of 5 simple, multi-layer perceptrons with two hidden layers of $100$ neurons and a dropout rate of 0.5. They were all independently randomly initialized and trained on the same data, but with data points presented in a randomly shuffled order. For further specifications we refer to Appendix \ref{app:models}.

\subsubsection{Variational Autoencoder} 
For obtaining an indication of novelty we trained a variational autoencoder (VAE) \citep{vae}. We picked this model because it is suitable for high-dimensional data and widely applicable to different data types. A VAE can be understood as a pair of neural networks, an encoder and a decoder. An encoder network maps an input to a lower dimensional latent distribution, while a decoder network attempts to convert this lower dimensional representation back to the original input. We use the VAE's reconstruction error as a measure of uncertainty, following the intuition that the autoencoder will only learn to reconstruct data that is similar to the data it has seen before. Once more, we refer to Appendix \ref{app:models} for further implementation details.
\subsubsection{MC Dropout}
For our implementation of MC Dropout \citep{gal2016dropout}, we train a single multi-layer perceptron with the same architecture and training procedure as described for the Deep ensemble members. During inference, $100$ forward passes are performed with the dropout layers activated and the average over these outputs is taken as a prediction. 

\subsubsection{Bootstrapped Logistic Regression}  The widely used method called `bootstrap' was introduced already in \cite{bootstrap}. It mimics repeated sampling from the true data distribution by resampling with replacement from the data set. For our present purpose, we created an ensemble by training multiple models on bootstrapped samples of the data set. We used this method as an intuitive simple counterpart to the deep ensemble, in order to investigate the influence and added value of model flexibility.  We again opted for an ensemble of 5 models, training each model on a different bootstrapped sample of the data of the same size as the training data. Again, we refer to Appendix \ref{app:models} for implementation details.
\\\\
For both ensemble models the prediction was calculated as the average over the predictions of the individual ensemble members. Uncertainty was calculated as the entropy of a prediction. This is an intuitive uncertainty estimate: it is highest when the predicted probability is $0.5$ and lowest when the probability is either $0$ or $1$.

\subsection{Evaluation methods}
\subsubsection{Metrics}

Alongside the area under ROC curve (abbreviated with AUC-ROC) to measure performance, 
we used the Expected Calibration Error (ECE) metric \citep{binningECE} to evaluate the calibration of our models. ECE is calculated as follows:
\begin{align*}
    ECE &= \frac{1}{N}\sum_{k=1}^KN_k|\bar{y}_k-\bar{o}_k|
\end{align*}
where $K$ is the number of mutually exclusive equally-sized bins,  $\bar{y}_k$ is the mean predicted probability within bin $k$ and $\bar{o}_k$ the observed outcome probability (fraction of positives) within $k$. In our implementation we employed 10 bins, so the first bin included all data points with a predicted probability between 0 and $10\%$, the second bin between $10\%$ and $20\%$, and so on. 

An uncertainty estimate is useful if the model can be trusted when it is confident, and can be ignored when it is uncertain.  Whether this is the case can be evaluated by sorting the predictions by their estimated uncertainty and tracking the performance metrics as larger and larger portions of uncertain data points are excluded, simulating what would happen to performance if we were to remove uncertain data points.

\subsubsection{Out Of Domain detection}
It is likely that a model will not give reliable predictions for patients that are not well represented in the training data. Therefore, a good uncertainty estimate should be higher for these patients. Out Of Domain (OOD) detection is often evaluated in computer vision, where one can simply feed a picture from another domain to a classifier and check whether the model is indeed uncertain. Alas, when dealing with tabular data this is not as straightforward. We devised a number of scenarios to simulate clinically-relevant situations
and evaluated the uncertainty estimate empirically. The evaluation was performed by excluding a specific group of patients from the training data and then testing the model on this excluded group as well as on a normal test set without patients from this group. We expect the uncertainties to be higher on this OOD group as compared to the normal test set, which consists of patients similar to the training data. To quantitatively evaluate the ability of detecting these OOD groups based on uncertainty, we followed \cite{hendrycks17baseline}: all predictions, both from the normal test set and the OOD group, were ranked according to estimated uncertainty, thresholds were used to predict which samples are OOD and AUC-ROC was calculated for the varying threshold. As a consequence, AUC-ROC is highest if all OOD patients have higher uncertainty than all patients in the regular test set. 
We experimented on the following scenarios:\footnote{See Appendix \ref{app:OOD} for the descriptives of the various groups. }
\paragraph{Rare or new disease} One possibility is that a clinic or a ward starts receiving patients with an illness not previously treated in that structure, either because of a new pathology or because of a change in protocol. We simulate this by using ICD-9 diagnoses that are given at the end of each hospital stay, labeling as OOD all patients with a specific pathology.
\paragraph{Generalization to other ethnic groups and genders} We could imagine that a certain population group, such as an ethnic minority or gender, is not represented in the training data. In this context we would also want a higher uncertainty for such OOD patients to avoid unwarranted generalizations.           
\paragraph{Different admission type} Another scenario could be a change in hospital organisation leading to a different patient distribution. For example, a new procedure could get instated where a specific type of patients (e.g. a specific type of surgery) now gets referred to the Intensive Care, whereas previously they were referred to another ward. We simulate this situation by considering as OOD either the emergency/urgent patients (unplanned medical care), or the elective patients (planned hospital admission).

\paragraph{Corrupted data} Another OOD group consists of patients with abnormal feature values, for example, arising as a consequence of faulty sensors, software bugs or human typing errors. To simulate this mishaps we construct a perturbed version of the test set by picking a random feature and re-scaling it across the test set.  The desired outcome is higher uncertainties for the perturbed test set compared to the regular test set. We repeat the same procedure for 30 randomly selected features, as different features might have different effects, and then average the results. This perturbation experiment is repeated twice with a multiplication factor of 10 and 1000.

\section{Cohort}
We performed experiments on the MIMIC-III data set \citep{MIMIC}, which comprises health data from IC unit admissions from the Beth Israel Deaconess Medical Center in Boston, Massachusetts.  MIMIC-III is well-documented and represented in literature with benchmarks for comparison \citep{mimicbenchmark2, benchmarking}. 
We used the same data extraction pipeline as \cite{benchmarking}, who specified multiple benchmark tasks. We focused only on risk prediction  of in-hospital mortality based on the first 48 hours of data. From the $46476$ patients and $61532$ IC stays in the MIMIC-III data set, data points are selected based on patient age ($\geq18$ years) and length of stay ($\geq48$ hours). Furthermore, stays were excluded when there was no data in the first 48 hours or when there were multiple IC transfers within one hospital admission. After selection, the cohort consisted of $21139$ IC stays corresponding to $18094$ patients, with a mortality rate of $13.23\%$. 
We employed the feature engineering for the Logistic Regression model in \citep{benchmarking}, where for 17 selected clinical variables, six statistics are calculated on seven sub-sequences of a time series. The sub-sequences consist of the full time series, the first 10\% and last 10\%, the first and last 25\%, and the first and last 50\%. The statistics are minimum, maximum, mean, standard deviation, skew and the number of measurements,  which are all standard-scaled. This results in 714 features in total.
We also adopted the same train, test and validation split, of size 14681, 3236, 3222 respectively. 
Additionally, we used toy data to highlight key intuitions. For a description of the toy data generation process we refer to Appendix \ref{app:toy}. 

\paragraph{Churn data}
Additionally, we perform experiments on another data set\footnote{The data set can be found here: \texttt{https://www.kaggle.com/shrutimechlearn/churn-modelling}},  which contains data of bank customers. The binary classification task is to predict whether the customer left the bank or not. Even though it is not a medical data set, we use it because it has similar characteristics: an unbalanced target and a combination of categorical and continuous features. 
We apply standard scaling to the 13 features and perform a random split in train, test and validation set, of size 6000, 2000, and 2000 respectively.

\section{Results} 
The desired properties of an uncertainty estimate are correlation between performance and uncertainty, and effective detection of OOD. To understand the behaviour of different methods on high-dimensional data, we first conduct experiments on 2-dimensional toy data.

\subsubsection{What to expect: simulation on toy data}
\label{subsec:toyresults}

\begin{figure}[ht]
  \subfloat[]{
	\begin{minipage}[c][1\width]{
	   0.32\textwidth}
	   \centering
	   \includegraphics[width=1\linewidth]{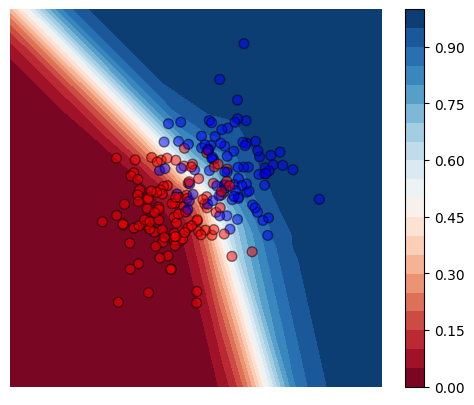}
	   \label{fig:nn_single}
	\end{minipage}}
 \hfill 	
  \subfloat[]{
	\begin{minipage}[c][1\width]{
	   0.32\textwidth}
	   \centering
	   \includegraphics[width=1\linewidth]{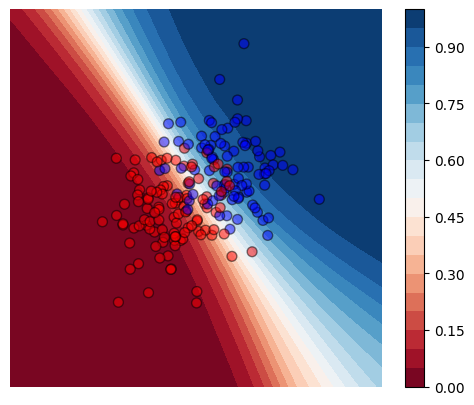}
	   \label{fig:nn_ensemble}
	\end{minipage}}
	\hrulefill
	  \subfloat[]{
	\begin{minipage}[c][1\width]{
	   0.32\textwidth}
	   \centering
	   \includegraphics[width=1\linewidth]{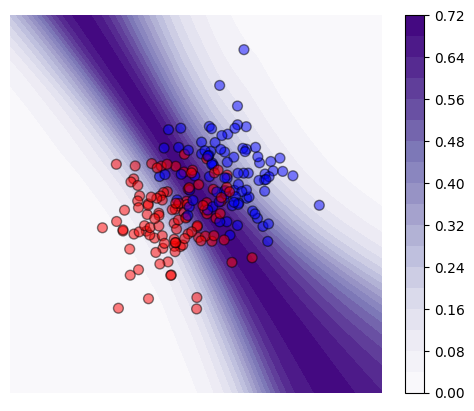}
	   \label{fig:nn_ensemble_entr}
	\end{minipage}}
\caption{Showing the predicted probabilities of a single neural network model (a), the average of the predicted probabilities of differently initialized neural networks  (b) and the entropy of these predicted probabilities (c).}
\label{fig:nn_example}
\end{figure}

We showcase what the different methods capture on toy data. The red and blue points denote data points from two different classes. In Figure \ref{fig:nn_example} we observe the difference between using a single model and an average over ensemble members: the ensembles produce a wider decision boundary further away from the data, reflecting the diverging beliefs of different single models. In Figure\ref{fig:nn_ensemble_entr} we show how the predicted probabilities translate to uncertainty, defined by the entropy. 
We can observe similar behaviour with Bootstrapped LR and MC Dropout, for which the same plots are shown in Appendix \ref{app:extratoy}.

\begin{figure}[ht]
  \subfloat[]{
	\begin{minipage}[c][1\width]{
	   0.32\textwidth}
	   \centering
	   \includegraphics[width=1\linewidth]{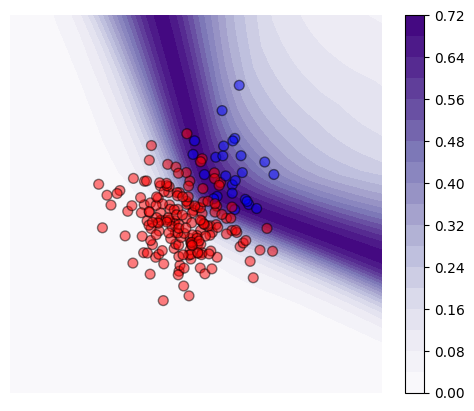}
	   \label{fig:unb_noweighting}
	\end{minipage}}
 \hfill 	
  \subfloat[]{
	\begin{minipage}[c][1\width]{
	   0.32\textwidth}
	   \centering
	   \includegraphics[width=1\linewidth]{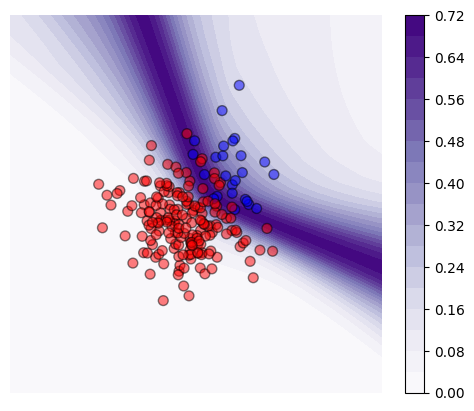}
	   \label{fig:unb_weighting}
	\end{minipage}}
	\hfill 	
  \subfloat[]{
	\begin{minipage}[c][1\width]{
	   0.32\textwidth}
	   \centering
	   \includegraphics[width=1\linewidth]{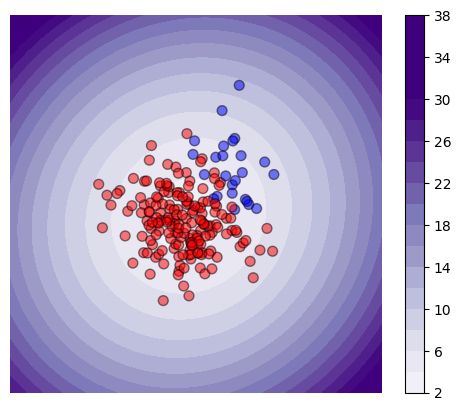}
	   \label{fig:unb_gmm}
	\end{minipage}}
\caption{Showing the uncertainties of an ensemble of neural networks with (a) and without using class weighting (b). In (c) we plot the estimated negative log likelihood to illustrate how the minority class might lie in a low-density area.}
\label{fig:classweighting}
\end{figure}

In Figure \ref{fig:classweighting} we see an example of what might happen in an unbalanced setting. The model focuses most on learning to predict the majority class with high certainty. As a consequence, when using entropy to reject examples based on uncertainty we will reject a large fraction of positive examples. A way to solve this problem, also employed in \cite{leibig2017leveraging}, is using class weighting in the loss, i.e. artificially putting more weight on the minority class\footnote{The exact implementation of class weighting is described in Appendix \ref{app:models}.}. This results in more confident predictions for the most obvious minority data points. Figure \ref{fig:unb_gmm} showcases that, in an imbalanced problem, the minority points might lie in a relatively low-density area of the space.

These plots (together with those reported in Appendix \ref{app:extratoy}) clearly illustrate what we expect to see on high-dimensional data. Single models, ensembles and MC dropout have a clear correlation between uncertainty and decision boundary, but their area of certainty tends to generalize to vast unseen areas of the feature space: we expect them to have good confidence-performance plots and to struggle with OOD detection. Conversely, density (or novelty) estimation should detect OOD more easily but also assign high certainty to points in a dense are of overlap between classes, hindering the correlation between confidence and discrimination.

\subsubsection{Confidence-performance plots on MIMIC-III}
We turn to the experiments on real-world data, for the prediction of mortality in IC. We show the confidence-performance plots with and without using class weighting. We employ entropy of the prediction as the uncertainty estimate, both for the single model and the ensemble model (where entropy of the average prediction 
To evaluate the VAE uncertainties, we use the single neural network as our classifier, but reject patients based on VAE reconstruction errors.  

\begin{figure}[ht]
  \subfloat[]{
	\begin{minipage}[c][1\width]{
	   0.32\textwidth}
	   \centering
	   \includegraphics[width=1\linewidth]{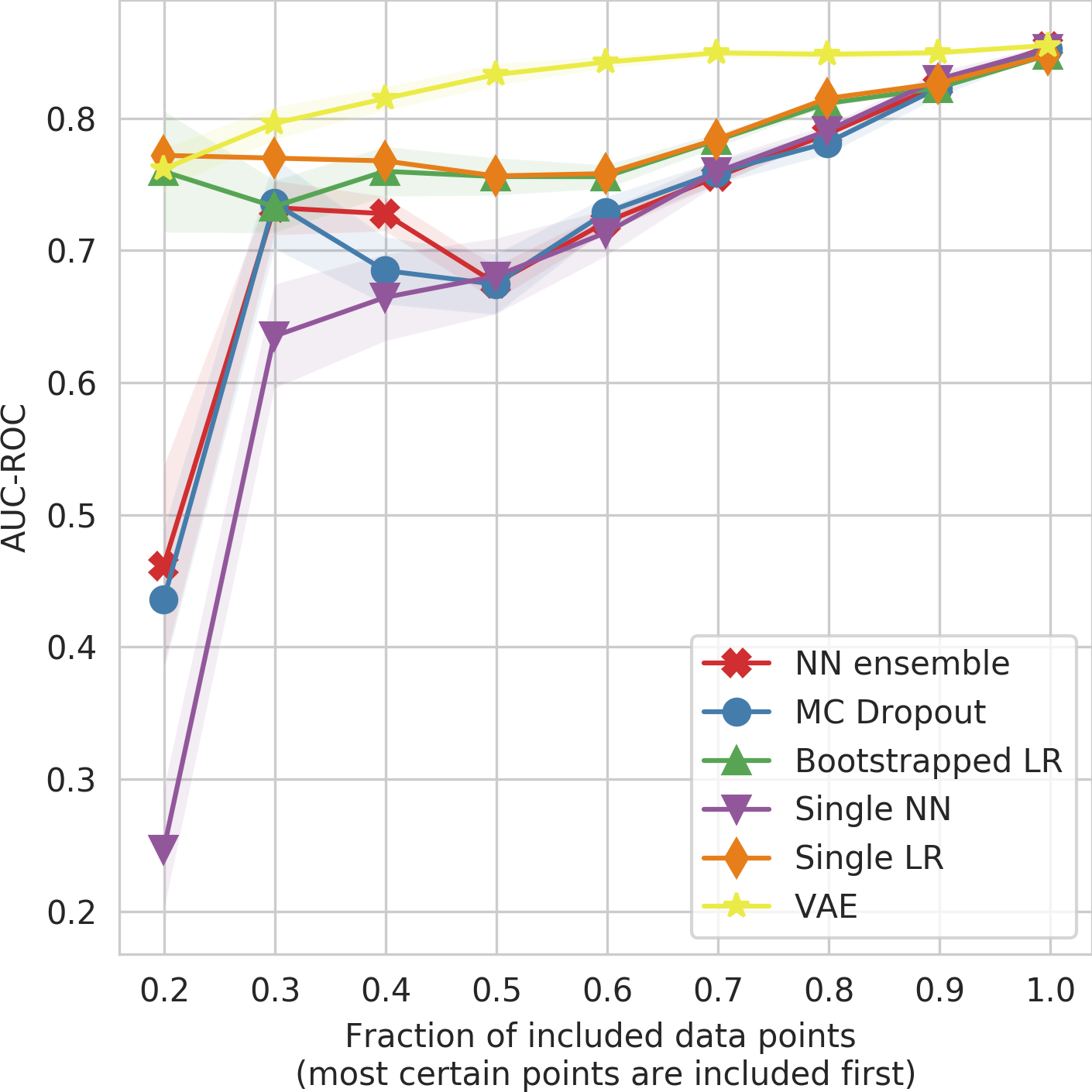}
	   \label{fig:aucs_noclassweight}
	\end{minipage}}
 \hfill 	
  \subfloat[]{
	\begin{minipage}[c][1\width]{
	   0.32\textwidth}
	   \centering
	   \includegraphics[width=1\linewidth]{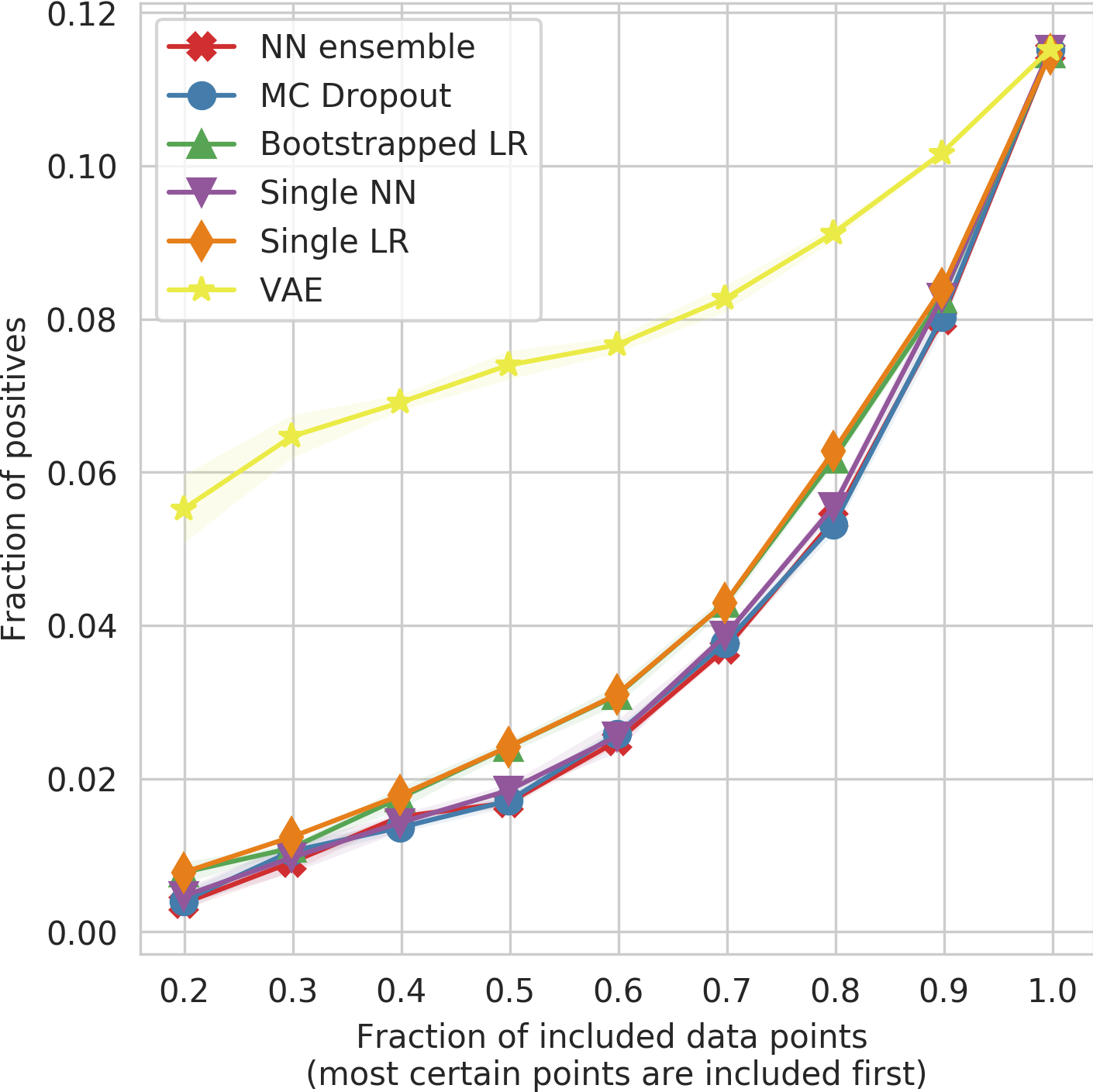}
	   \label{fig:avgy_noclassweight}
	\end{minipage}}
	\hrulefill
	  \subfloat[]{
	\begin{minipage}[c][1\width]{
	   0.32\textwidth}
	   \centering
	   \includegraphics[width=1\linewidth]{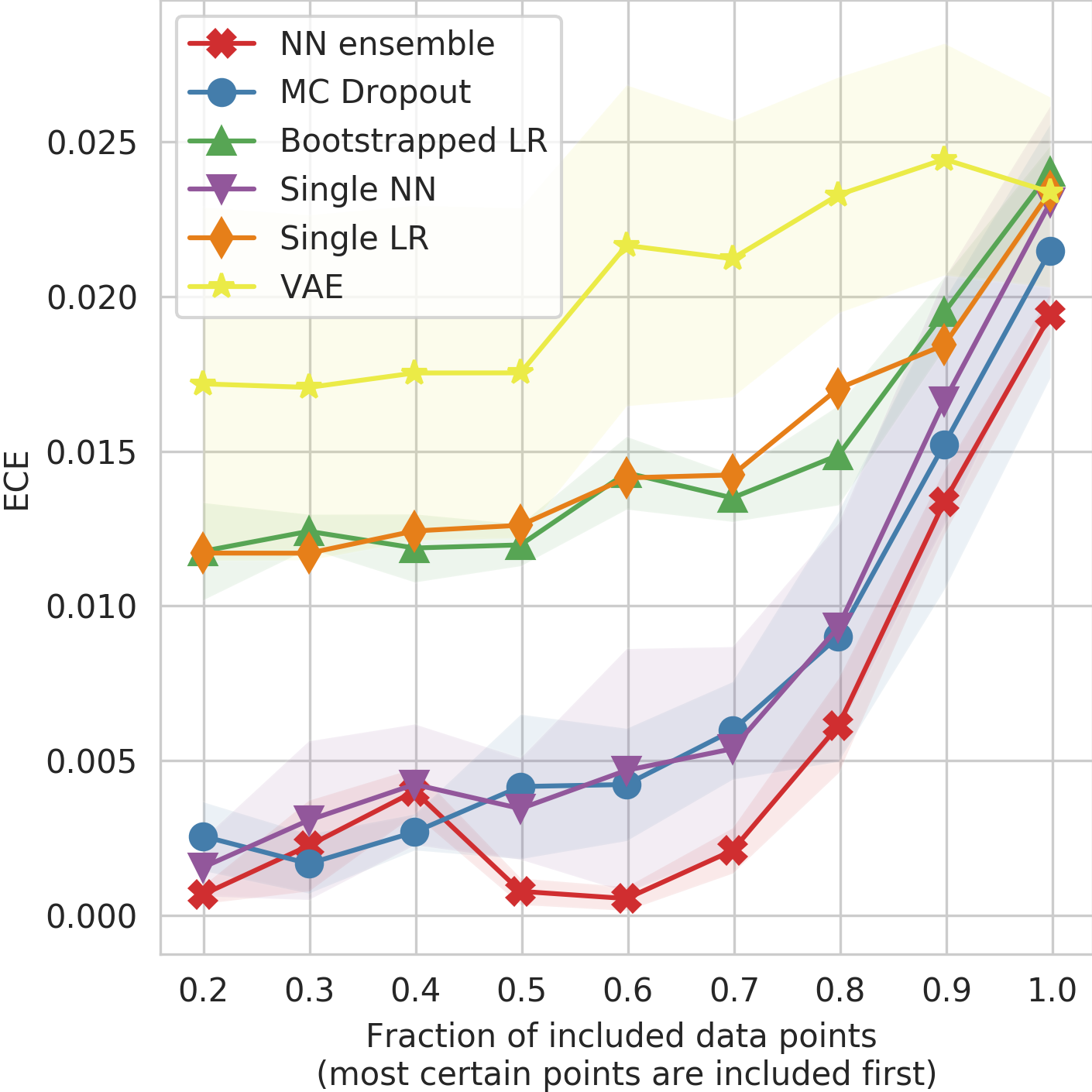}
	   \label{fig:ece_noclassweight}
	\end{minipage}}
\caption{Confidence-performance plots without using class weighting. Showing mean and standard deviation over 5 random seeds.}
\label{fig:conf-perf-noweights}
\end{figure}

\paragraph{Without class weighting} Figure \ref{fig:aucs_noclassweight} shows AUC-ROC against the fraction of included test data points, where the most certain points are included first.  We observe that for all methods AUC-ROC worsens when excluding uncertain points. This is counter-intuitive but can be explained by class imbalance: since the classes are unbalanced, the model put most focus on learning to predict the majority class (in this case the negative class) with high certainty, like in Figure \ref{fig:unb_noweighting}. 

The fact that the most certain points are points from the majority class, can be seen in Figure \ref{fig:avgy_noclassweight}, where the fraction of positive examples in the confident subset is shown. In Figure \ref{fig:ece_noclassweight} we observe that the calibration error is very low in all subsets, and seems to be smaller for the most certain data points. This is desirable, but the differences are so small that it is hard to say whether they are true effects or artifacts of binning. 

\begin{figure}[ht]
  \subfloat[]{
	\begin{minipage}[c][1\width]{
	   0.32\textwidth}
	   \centering
	   \includegraphics[width=1\linewidth]{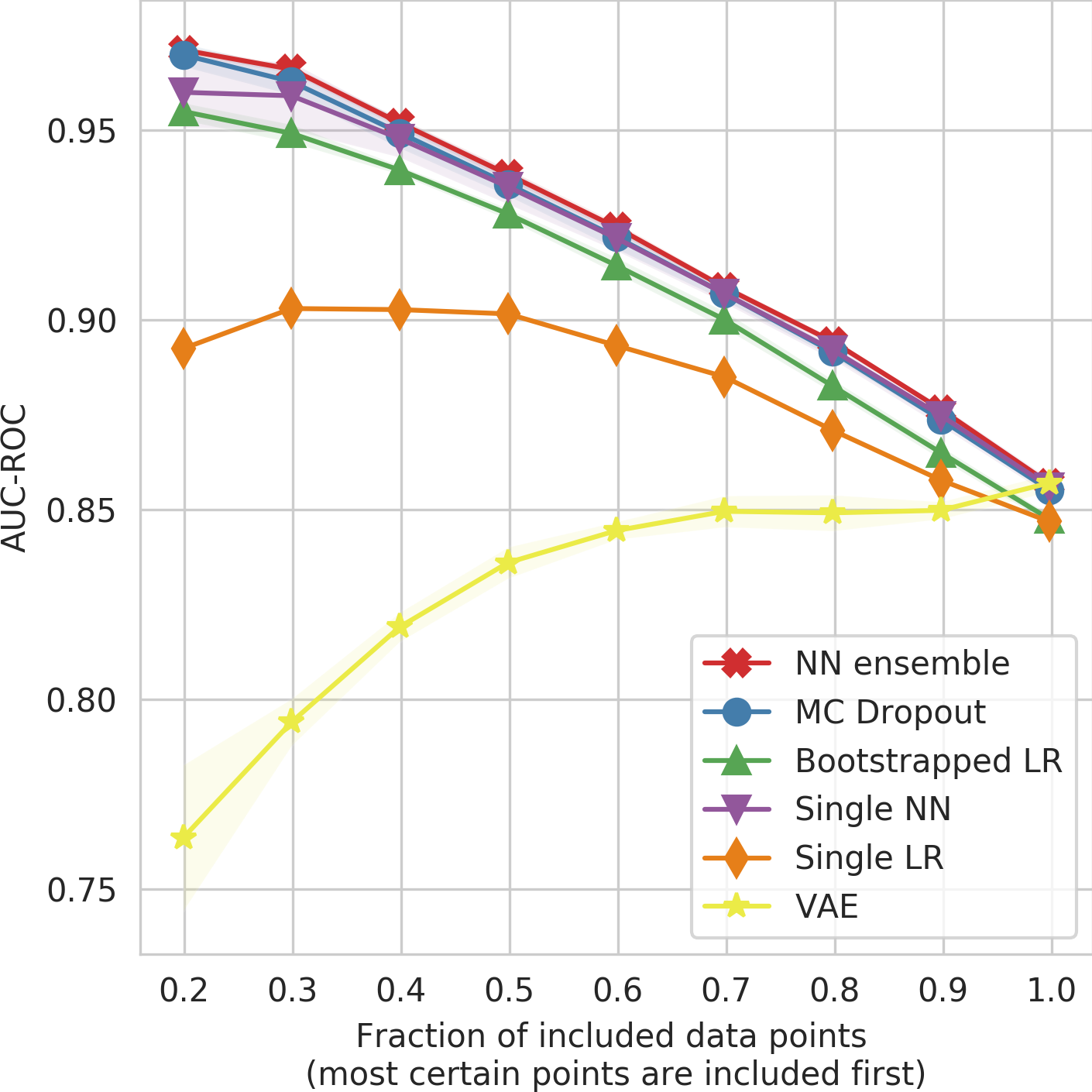}
	   \label{fig:aucs_classweight}
	\end{minipage}}
 \hfill 	
  \subfloat[]{
	\begin{minipage}[c][1\width]{
	   0.32\textwidth}
	   \centering
	   \includegraphics[width=1\linewidth]{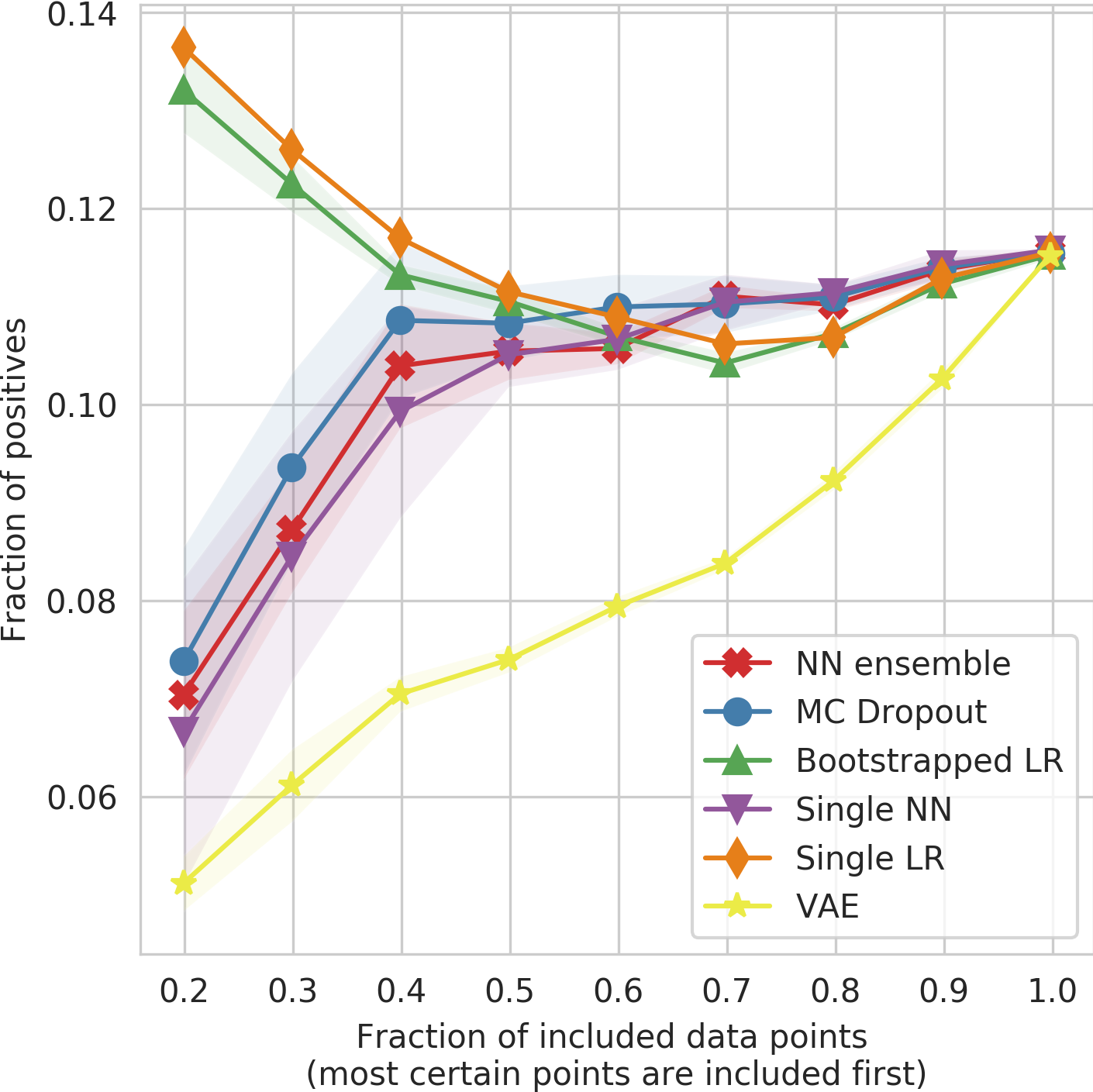}
	   \label{fig:avgy_classweight}
	\end{minipage}}
	 \hfill 	
  \subfloat[]{
	\begin{minipage}[c][1\width]{
	   0.32\textwidth}
	   \centering
	   \includegraphics[width=1\linewidth]{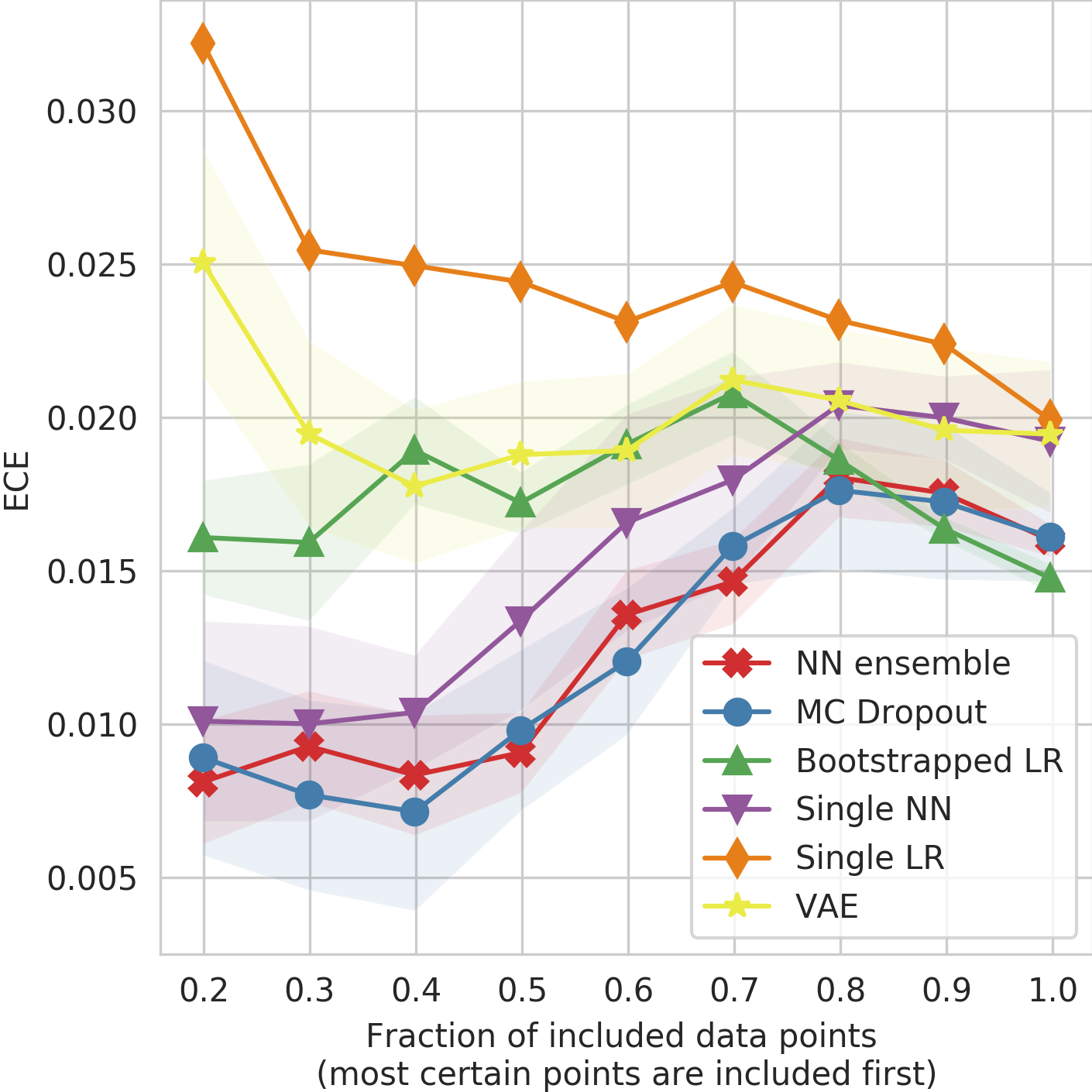}
	   \label{fig:ece_classweight}
	\end{minipage}}
\caption{Confidence-performance plots on the MIMIC data set when using class weighting. Showing mean and standard deviation over 5 random seeds.}
\label{fig:conf-perf}
\end{figure}

\paragraph{With class weighting} When using class weighting, the models artificially devote more attention to the minority class. We see this back in Figure \ref{fig:conf-perf}: the most confident data points do not only originate from the majority class but are relatively balanced. We also observe in Figure \ref{fig:aucs_classweight} that AUC-ROC now improves when excluding the most uncertain data points. For both LR and the NN, the ensemble versions have a similar AUC on all data as compared to the single models. However, the uncertainty estimates of the ensemble models seem to be better as they lead to more improvement in AUC-ROC when excluding uncertain data points. This suggests that the ensemble methods capture some measure of epistemic uncertainty on top of the aleatory uncertainty captured by the single models. Nevertheless, when looking from a practical point of view, it is unlikely that we would want to use uncertainty to exclude more than $20\%$ of the data. Keeping this in mind, we see that the difference in AUC-ROC between the ensemble methods and the single methods is relatively small when excluding only $20\%$.  This means that it might be a reasonable choice to use the baseline uncertainty in practice: especially the single NN has well-calibrated uncertainty estimates.

Interestingly,
excluding uncertain predictions based on the VAE uncertainty results in a slightly lower AUC-ROC. AUC-ROC likely decreases because some of the most obvious positive cases are in a low density area, e.g. the extremely sick. We see this in Figure \ref{fig:avgy_noclassweight}, where the percentage of positive points is lower for the most confident subsets of the data.

Finally, class weighting causes the risk predictions to be poorly calibrated. Therefore, to be interpreted as true risk, predictions need to be corrected. An easy and effective way to re-calibrate predictions is by Platt scaling via a held-out validation set (\cite{platt1999probabilistic}). We see in Figure \ref{fig:ece_classweight} that after this additional step, we can still obtain a good calibration on the different confident subsets of the data, with an expected calibration error of around 0.02. 

\begin{figure}[ht]
  \subfloat[]{
	\begin{minipage}[c][1\width]{
	   0.32\textwidth}
	   \centering
	   \includegraphics[width=1\linewidth]{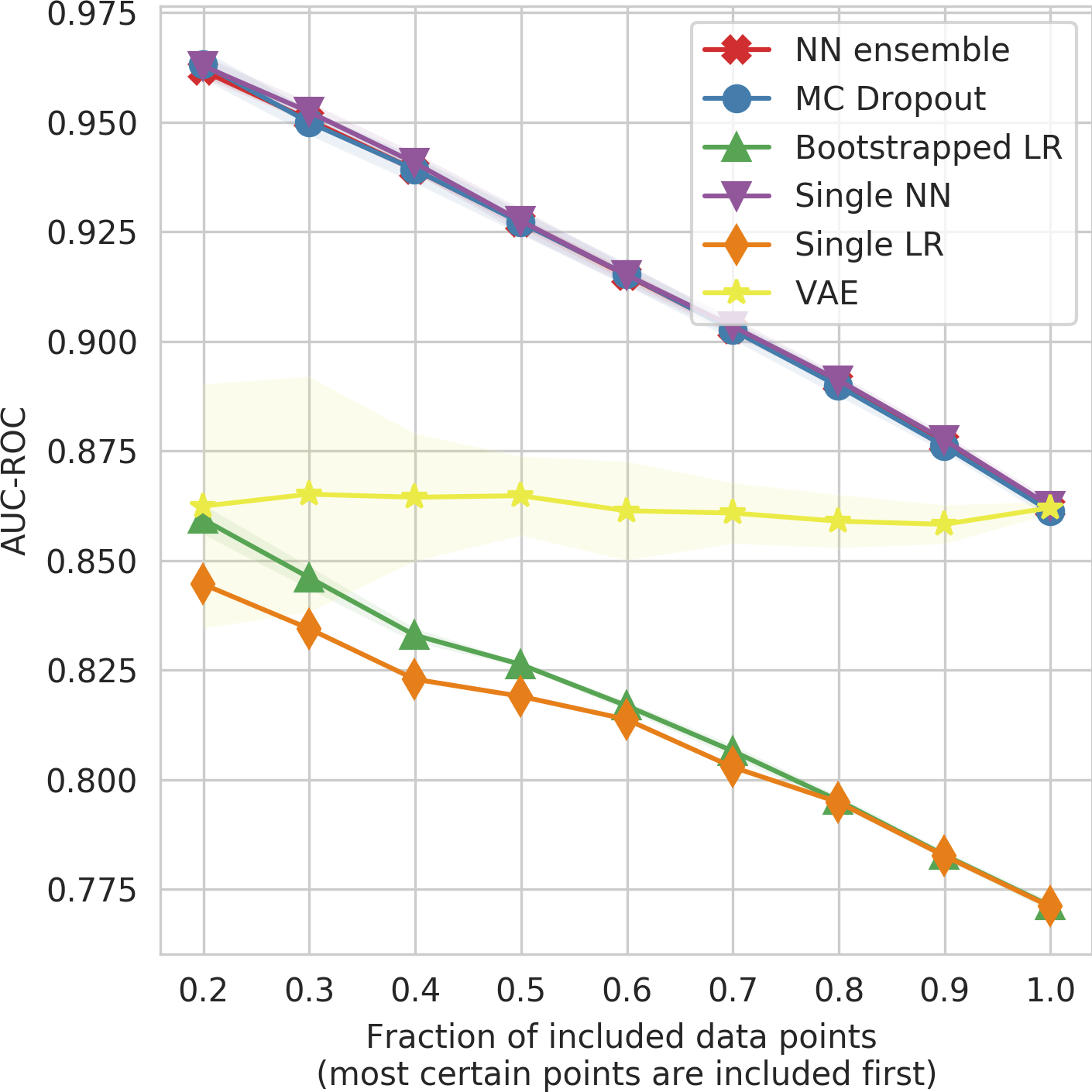}
	   \label{fig:aucs_classweight-churn}
	\end{minipage}}
 \hfill 	
  \subfloat[]{
	\begin{minipage}[c][1\width]{
	   0.32\textwidth}
	   \centering
	   \includegraphics[width=1\linewidth]{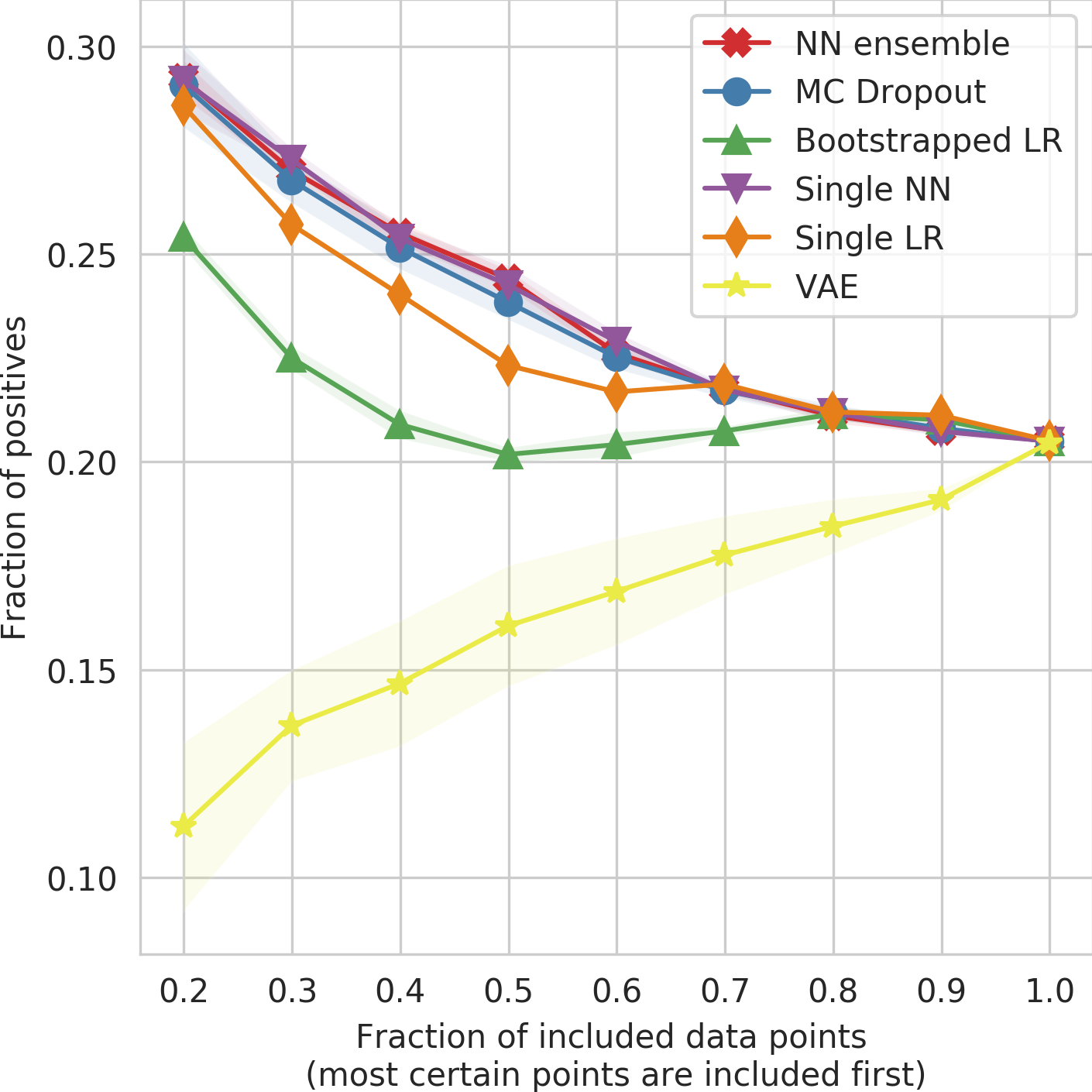}
	   \label{fig:avgy_classweight-churn}
	\end{minipage}}
	 \hfill 	
  \subfloat[]{
	\begin{minipage}[c][1\width]{
	   0.32\textwidth}
	   \centering
	   \includegraphics[width=1\linewidth]{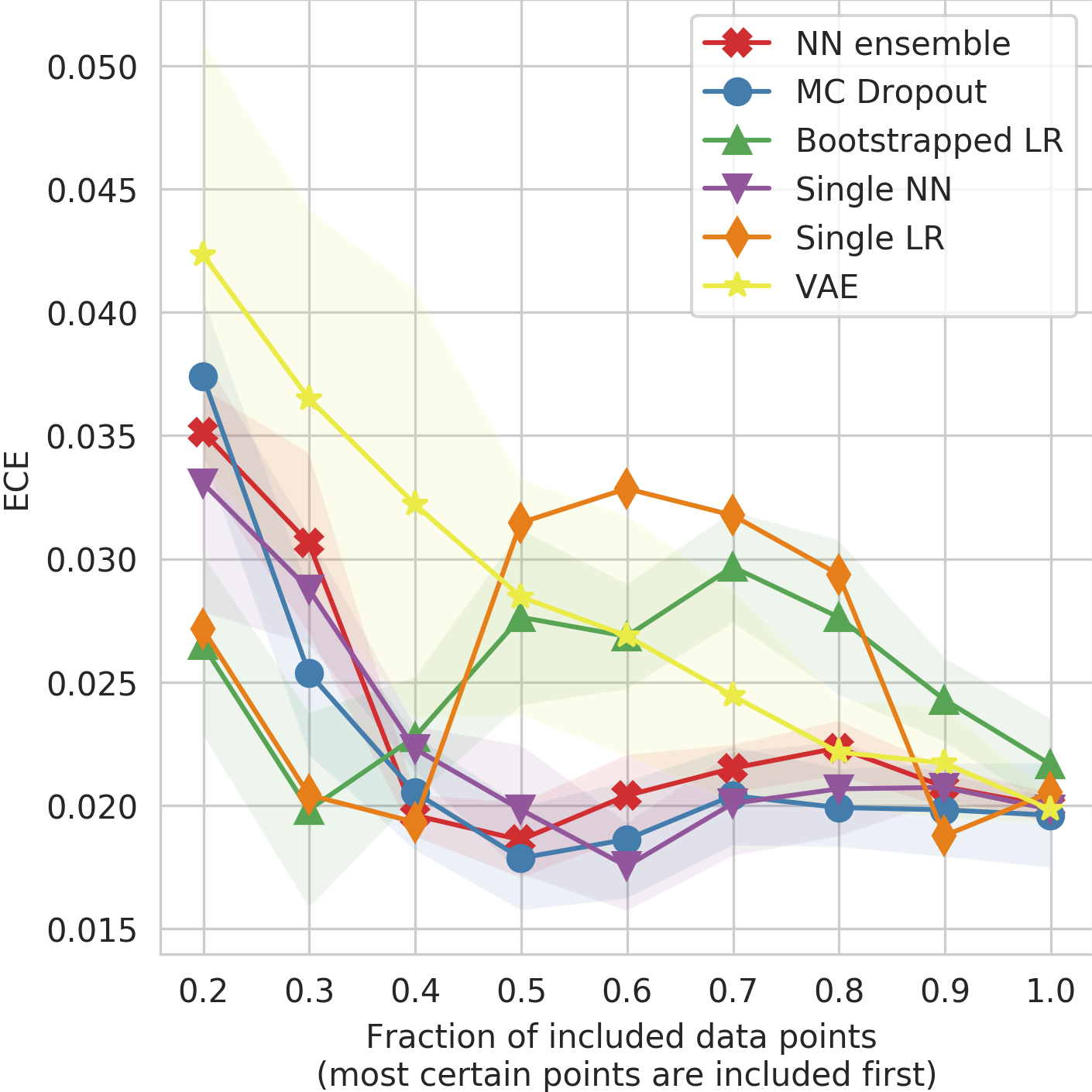}
	   \label{fig:ece_classweight-churn}
	\end{minipage}}
\caption{Confidence-performance plots on the Churn data set when using class weighting. Showing mean and standard deviation over 5 random seeds.}
\label{fig:conf-perf-churn}
\end{figure}

We observe very similar results on the Churn data, as shown in Figure \ref{fig:conf-perf-churn}. One noticeable difference is the confidence-performance of the VAE, which remains stable, indicating that the NN discriminates equally well on dense and sparse areas.

\label{app:churn}


\subsubsection{OOD experiment}
\begin{figure}
    \centering
    \includegraphics[width=1\linewidth]{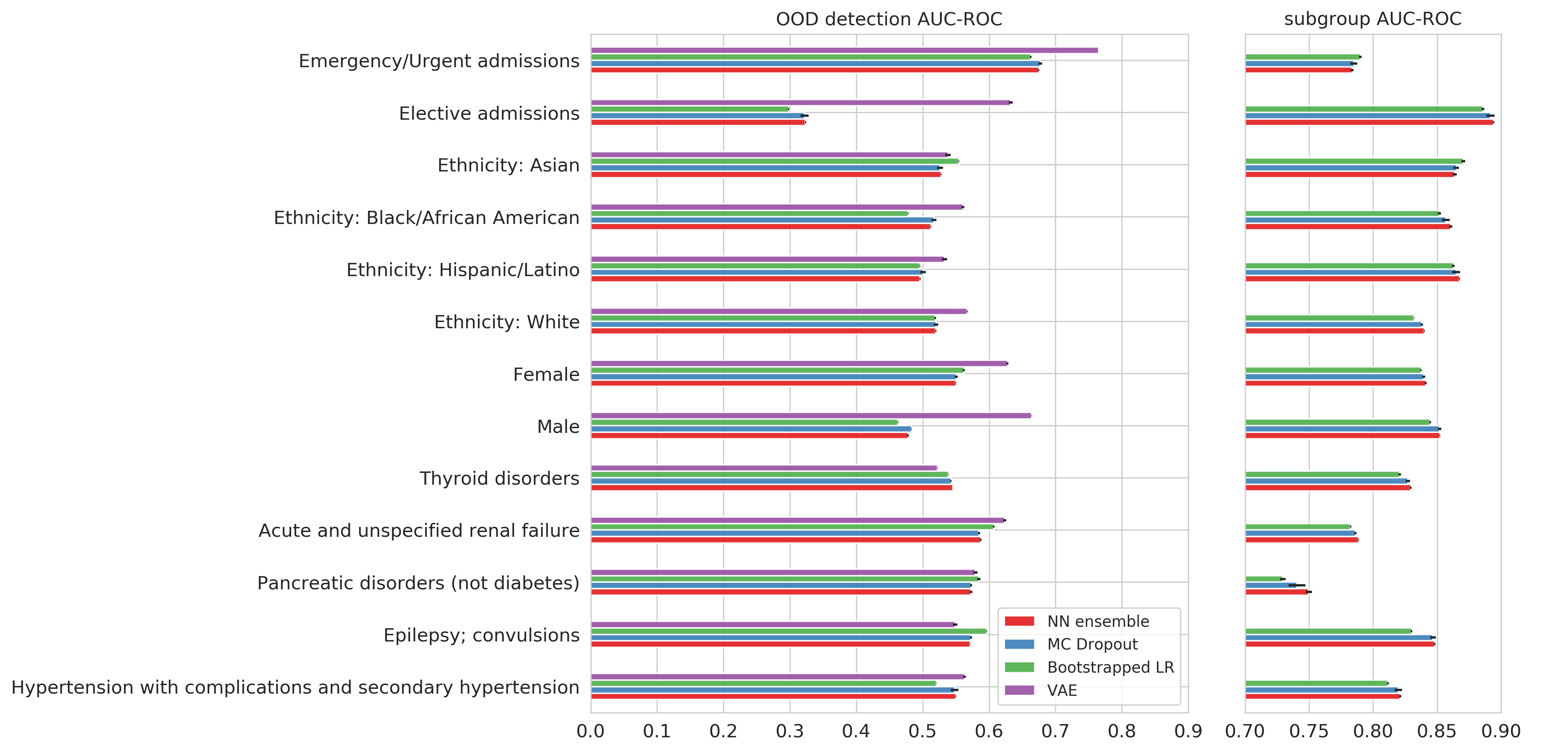}
    \caption{OOD detection and AUC-ROC of the model on the corresponding subgroup.}
    \label{fig:ood}
\end{figure}

\paragraph{Detection of OOD}In Figure \ref{fig:ood} we see the AUC-ROC scores of OOD detection using the uncertainty of the different class-weighted ensemble methods compared to the VAE. We observe that using the reconstruction error of the VAE,  each OOD group is detected with an AUC-ROC higher than $0.5$, indicating that the characteristics of the OOD patients are indeed (at least slightly) different from the training data. On the contrary, bootstrapped LR, NN ensemble and MC Dropout do not seem to detect the OOD groups effectively. The group that is most clearly detected is the group of emergency admissions, while the group of elective admissions is not recognized as OOD and receives a lower uncertainty than the test set. The other groups all have an AUC-ROC close to $0.5$, meaning the uncertainties are equally high for the OOD group compared to the test set. 

\paragraph{Subgroup AUC-ROC} As an additional check, we contrast OOD detection with generalization to unseen subgroups. We train a model on all data except the selected OOD patients, and evaluate the trained model on these OOD patients. This gives an idea of how well the model can generalize to this unseen group: if the model generalizes  well, then arguably it does not matter if the predictions are overly confident; if on the other hand the model does not generalize well, we would want to be warned for this by high uncertainty. In general, we see in Figure \ref{fig:ood} that this seems to be the case. Consider the model that had not seen any emergency admission in the training data, on the first line. This model does not generalize very well to emergency admissions (with a relatively low AUC-ROC), but the same group also receives higher uncertainty estimates, so the model seems to know what it does not know. For the group of elective admissions we see an opposite phenomenon: the predictions are relatively confident even though the model has not seen elective admissions. However, the performance on this subgroup is still very high. This pattern seems to repeat in most subgroups, where better OOD detection (higher uncertainty) relates to worse subgroup performance and low uncertainty relates to better performance. 

To conclude, ensemble models do not generally give higher uncertainty to patient groups that are different from the training data, contrary to the VAE. However, ensembles seem to give higher uncertainty to groups on which the classification ability does not generalize well, which is likely the most important aspect. 

\subsubsection{Corrupted data as OOD}
\begin{figure}[ht]
 \hfill
  \subfloat[]{
	\begin{minipage}[c][1\width]{
	   0.49\textwidth}
	   \centering
	   \includegraphics[width=1\linewidth]{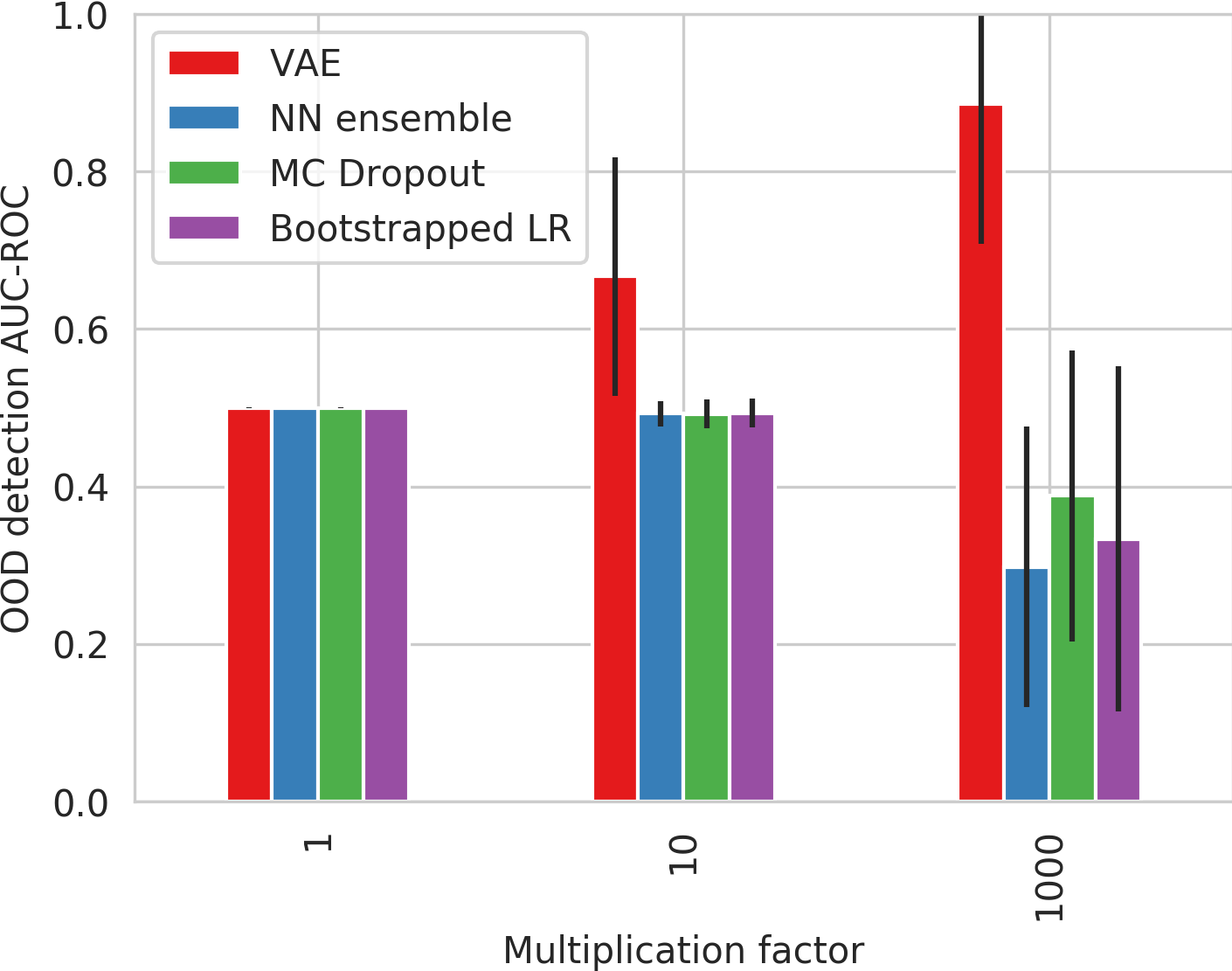}
	\end{minipage}}
  \hfill 	
  \subfloat[]{
	\begin{minipage}[c][1\width]{
	   0.49\textwidth}
	   \centering
	   \includegraphics[width=1\linewidth]{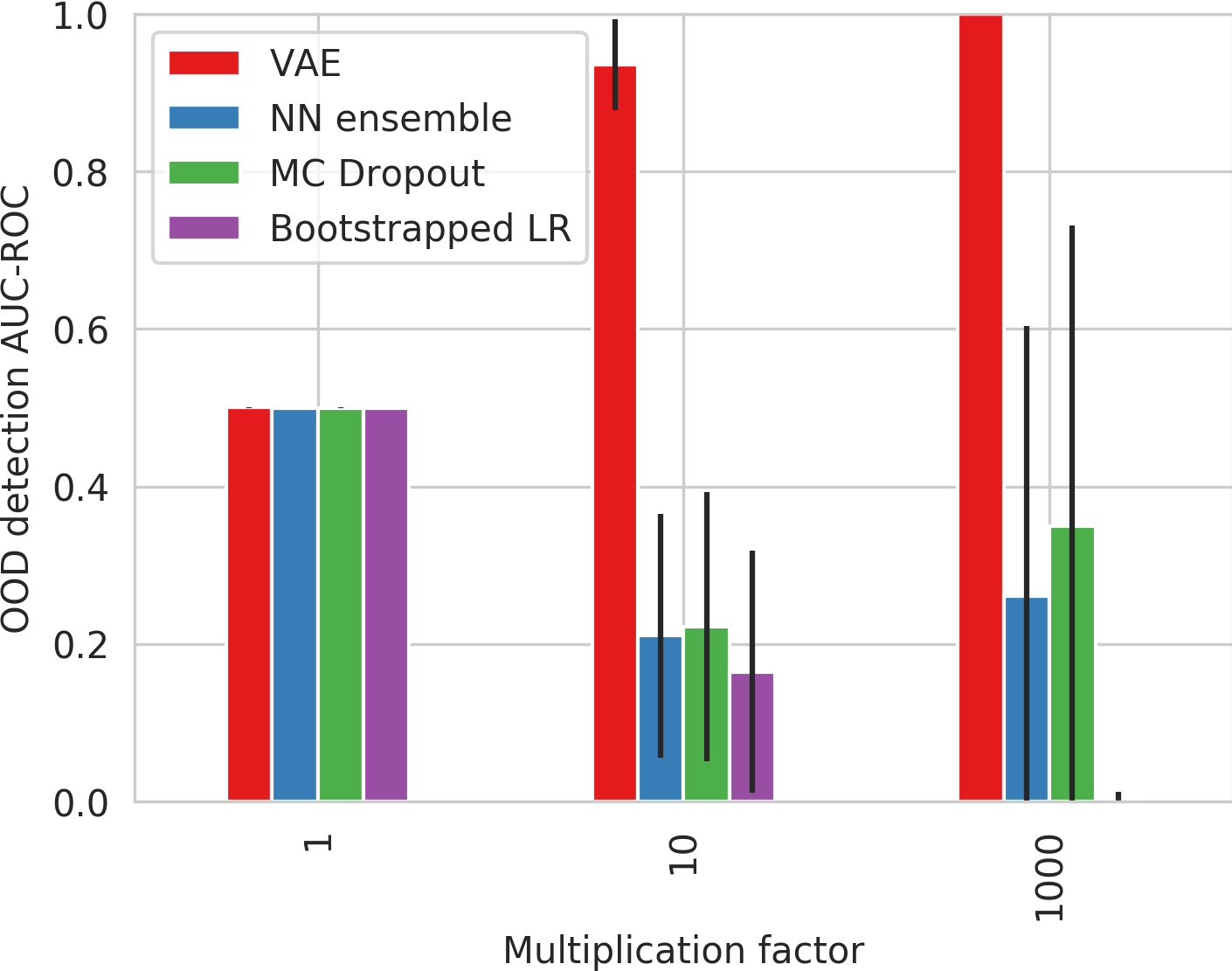}
	\end{minipage}}
\caption{The mean and standard deviation of the detection AUC-ROC over 30 different randomly re-scaled features, for three different scaling factors. In (a) the results on MIMIC-III, in (b) the results on the Churn data set. }
\label{fig:perturb}
\end{figure}

 We observe in Figure \ref{fig:perturb} that on real data the VAE gives on average higher uncertainty to corrupted data compared to regular test data, whereas the other models do not. Instead, ensemble models get on average \textit{more certain} when re-scaling a random feature, thus resulting in worse AUC-ROC for the task of detecting OOD. Notably, on the Churn data set the Bootstrapped LR has an extremely low AUC-ROC due to the lack of flexibility of said model.

 This behaviour is explained by the experiments on toy data in Figure \ref{fig:classweighting} and Figure \ref{fig:nn_ensemble_entr}: if we move very far from the training data, as we simulate in the corrupted data scenario, the ensemble model can still be very confident.  
Even giving the model more flexibility or decreasing the data set size did not change the fact that there are always confident areas far away from the training data. 

In light of the task at hand, this behaviour could be reasonable. For example, consider a situation where a model learns a linear relation between body temperature and risk. Even though a body temperature of $100\ ^o C$ does not exist, it is reasonable for the model to think that the patient has an extremely high risk. This underlines a tendency to generalize that is both a limitation and a strength of the task-specific uncertainty methods.

\section{Discussion} 

We summarize here our findings.
\begin{itemize}
\item When it comes to confidence-performance correlation,  uncertainty estimates belonging to the parameter uncertainty family (ensembles and MC Dropout) perform as desired, while the VAE depends on the performance of the associated classifier on more or less dense area, as well as the density of the area where classes inherently overlap.

\item On the clinically-informed experiments treating specific subgroups as OOD, the VAE performed remarkably better than other methods on gender and ethnicity subgroups as well as emergency/elective admissions. All methods seemed more or less on par when it came to detecting groups of patients with specific new illnesses. This suggests that detection of unforeseen diseases might be a especially challenging problem for medical AI employing tabular data.

\item In the corrupted data experiment the gap between the VAE and other methods widened, with the VAE being the only method able to reliably detect corrupted data. Other methods performed very poorly, a results that is at odds with the findings about ensembles in \citep{snoek2019can} and underscore the need for thorough testing on tabular data.
\end{itemize}

\paragraph{Limitations} The validity of these observations is conditional on the scarce availability of public medical tabular data: a certain degree of confidence can be obtained by using similar data but more testing is required on clinical datasets. Furthermore, our conclusions are limited to the methods included in our shortlist, and other approaches would require further experimentation. Finally, the added value of uncertainty estimation in medical AI should ultimately be assessed via real-world validations.
\\\\
In light of our findings, we draw some conclusions on the best usage of the aforementioned uncertainty techniques for two use cases -  classification and risk prediction.

\subsection{Classification}
In a classification setting, we are interested in predicting hard labels with confidence. The clearest role of uncertainty is to abstain from giving predictions for patients for which the label is uncertain.

Therefore, the most important property of an uncertainty estimate in this setting would be improved discrimination (measured by AUC-ROC) when excluding uncertain predictions. Furthermore, a high uncertainty on OOD patient groups is desirable. Finally, a secondary requirement would be that the predictions are well-calibrated. This is not necessary for the classification task, but could be useful if we are to interpret the risk predictions as probabilities to set a threshold. 

We have observed that by excluding predictions based on the confidence score of a probabilistic classifier, discrimination can significantly be improved. Using an ensemble of probabilistic classifiers has an even larger positive influence on AUC-ROC, for the NN ensemble but even for the simple LR ensemble.
It is worth remarking that this added value will be different dependent on data set size, model type and complexity of the data set. In some situations it might be worth the extra complexity and reduced interpretability, whereas in other cases it might not be necessary. 

Although ensembling methods result in larger improvement of AUC-ROC, these methods do not necessarily give higher uncertainty to OOD groups and could even be more confident about such data. 
The simulation of corrupted data taught us that we cannot expect to get high uncertainty estimates on data with abnormal feature values. This might not be a problem, as we could arguably deal with those cases in a preprocessing stage. It is however important to keep in mind that the model might still be overconfident about its own generalizability, even when using ensembling methods. Another difficulty about this approach is that the quality of an uncertainty estimate is sensitive to modeling assumptions of the ensemble members like the model architecture, as well as parameters specific to the ensemble such as the bootstrap size.

\subsection{Risk prediction} 
Often risk estimates are used directly to assist non-binary decisions or to stratify patients into groups (e.g. high, medium or low risk) for which different actions hold. Clearly an optimal model still discriminates perfectly, but in this approach we want to accept uncertain predictions if they are the best we can get given the data:  aleatory uncertainty is something we \textit{choose} to model and on which we base our decisions. Instead of wanting to assess the confidence about the true label, we want to know if we can trust the risk prediction. Hence this is different from the classification setting, where we aim to eliminate any uncertainty about the outcome, both aleatory and epistemic.

In this setting, the quality of the risk predictions is what we care about most. The calibration error gives a \textit{global} idea of quality of probabilistic predictions. If we would observe that the calibration error decreases when excluding uncertain predictions, this would indicate that uncertainty relates to prediction quality. Alas, local mistakes can be evened out by binning, which makes it hard to evaluate the influence of uncertainty on prediction quality with the calibration error.

Again, an improved AUC-ROC could indicate improved ranking and therefore that the uncertain predictions were less reliable. However, the quality of a risk prediction is not necessarily related to good discrimination. For example, if most data points are in areas with inherent overlap, the model will likely give most reliable risk predictions to this area with high aleatory uncertainty. It is important that the AUC-ROC stays reasonably high when excluding uncertain predictions, as otherwise the model would not be useful, but it does not necessarily need to improve.

We observed in Figures \ref{fig:ece_noclassweight} and \ref{fig:ece_classweight} that the calibration error improves when excluding uncertain predictions based on the uncertainty given by the ensembling methods, and that the calibration error is always slightly lower when using ensembling as compared to using a single model. However, we have seen that the calibration error on the total group of patients is already low or the predictions can easily be calibrated to result in a low calibration error. As the differences between calibration on the different confident subgroups are small, and the calibration metric is inherently imperfect because of binning artifacts, we cannot draw relevant conclusions about the relation between uncertainty and the quality of predictions.
 
Lacking a clear winner on the confidence-performance criteria, the VAE seems to be the most viable option, to at least ensure good OOD detection. We saw that excluding uncertain predictions based on the VAE uncertainty might result in a slightly lower AUC-ROC. 
In practice, this might not be a problem in a decision support setting, as obvious cases would also be simple for clinicians to handle without the help of a model's prediction. In this context, an indication of dissimilarity from previously seen cases could already be informative enough.

A limitation of having a separate novelty detection component is that we get an extra, possibly `black box' model that needs to be explained to the user. Furthermore, novelty detection is not a trivial task on its own.  Finally, another impediment when using this type of uncertainty to detect unreliable risk predictions is that it is hard to estimate or quantify the added value of the uncertainty estimate. This could however be measured empirically via a real-world trial, relating the use of uncertainty to patient outcomes and user trust.

\bibliography{references}

\begin{thebibliography}{34}
\providecommand{\natexlab}[1]{#1}
\providecommand{\url}[1]{\texttt{#1}}
\expandafter\ifx\csname urlstyle\endcsname\relax
  \providecommand{\doi}[1]{doi: #1}\else
  \providecommand{\doi}{doi: \begingroup \urlstyle{rm}\Url}\fi

\bibitem[Keane and Topol(2018)]{keane2018eye}
Pearse~A Keane and Eric~J Topol.
\newblock With an eye to ai and autonomous diagnosis, 2018.

\bibitem[Damen et~al.(2016)Damen, Hooft, Schuit, Debray, Collins, Tzoulaki,
  Lassale, Siontis, Chiocchia, Roberts, et~al.]{damen2016prediction}
Johanna~AAG Damen, Lotty Hooft, Ewoud Schuit, Thomas~PA Debray, Gary~S Collins,
  Ioanna Tzoulaki, Camille~M Lassale, George~CM Siontis, Virginia Chiocchia,
  Corran Roberts, et~al.
\newblock Prediction models for cardiovascular disease risk in the general
  population: systematic review.
\newblock \emph{bmj}, 353:\penalty0 i2416, 2016.

\bibitem[Jaspers et~al.(2011)Jaspers, Smeulers, Vermeulen, and
  Peute]{jaspers2011effects}
Monique~WM Jaspers, Marian Smeulers, Hester Vermeulen, and Linda~W Peute.
\newblock Effects of clinical decision-support systems on practitioner
  performance and patient outcomes: a synthesis of high-quality systematic
  review findings.
\newblock \emph{Journal of the American Medical Informatics Association},
  18\penalty0 (3):\penalty0 327--334, 2011.

\bibitem[Mosca et~al.(2011)Mosca, Benjamin, Berra, Bezanson, Dolor,
  Lloyd-Jones, Newby, Pina, Roger, Shaw, et~al.]{mosca2011effectiveness}
Lori Mosca, Emelia~J Benjamin, Kathy Berra, Judy~L Bezanson, Rowena~J Dolor,
  Donald~M Lloyd-Jones, L~Kristin Newby, Ileana~L Pina, V{\'e}ronique~L Roger,
  Leslee~J Shaw, et~al.
\newblock Effectiveness-based guidelines for the prevention of cardiovascular
  disease in women—2011 update: a guideline from the american heart
  association.
\newblock \emph{Journal of the American College of Cardiology}, 57\penalty0
  (12):\penalty0 1404--1423, 2011.

\bibitem[Geifman and El-Yaniv(2017)]{geifman2017selective}
Yonatan Geifman and Ran El-Yaniv.
\newblock Selective classification for deep neural networks.
\newblock In \emph{Advances in neural information processing systems}, pages
  4878--4887, 2017.

\bibitem[Der~Kiureghian and Ditlevsen(2009)]{der2009aleatory}
Armen Der~Kiureghian and Ove Ditlevsen.
\newblock Aleatory or epistemic? does it matter?
\newblock \emph{Structural Safety}, 31\penalty0 (2):\penalty0 105--112, 2009.

\bibitem[Urbina and Mahadevan(2011)]{urbina2011quantification}
Angel Urbina and Sankaran Mahadevan.
\newblock Quantification of aleatoric and epistemic uncertainty in
  computational models of complex systems.
\newblock In \emph{Structural Dynamics, Volume 3}, pages 519--535. Springer,
  2011.

\bibitem[Gal and Ghahramani(2016)]{gal2016dropout}
Yarin Gal and Zoubin Ghahramani.
\newblock Dropout as a bayesian approximation: Representing model uncertainty
  in deep learning.
\newblock In \emph{international conference on machine learning}, pages
  1050--1059, 2016.

\bibitem[H{\"u}llermeier and Waegeman(2019)]{uncertaintytutorial}
Eyke H{\"u}llermeier and Willem Waegeman.
\newblock Aleatoric and epistemic uncertainty in machine learning: A tutorial
  introduction.
\newblock \emph{arXiv preprint arXiv:1910.09457}, 2019.

\bibitem[Filos et~al.(2019{\natexlab{a}})Filos, Farquhar, Gomez, Rudner,
  Kenton, Smith, Alizadeh, de~Kroon, and Gal]{benchmarkingbayesian}
Angelos Filos, Sebastian Farquhar, Aidan~N. Gomez, Tim G.~J. Rudner, Zachary
  Kenton, Lewis Smith, Milad Alizadeh, Arnoud de~Kroon, and Yarin Gal.
\newblock Benchmarking bayesian deep learning with diabetic retinopathy
  diagnosis, 2019{\natexlab{a}}.

\bibitem[Leibig et~al.(2017)Leibig, Allken, Ayhan, Berens, and
  Wahl]{leibig2017leveraging}
Christian Leibig, Vaneeda Allken, Murat~Se{\c{c}}kin Ayhan, Philipp Berens, and
  Siegfried Wahl.
\newblock Leveraging uncertainty information from deep neural networks for
  disease detection.
\newblock \emph{Scientific reports}, 7\penalty0 (1):\penalty0 17816, 2017.

\bibitem[Ruhe et~al.(2019)Ruhe, Cina, Tonutti, de~Bruin, and
  Elbers]{ruhe2019bayesian}
David Ruhe, Giovanni Cina, Michele Tonutti, Daan de~Bruin, and Paul Elbers.
\newblock Bayesian modelling in practice: Using uncertainty to improve
  trustworthiness in medical applications.
\newblock \emph{arXiv preprint arXiv:1906.08619}, 2019.

\bibitem[Fumera and Roli(2004)]{fumera2004analysis}
Giorgio Fumera and Fabio Roli.
\newblock Analysis of error-reject trade-off in linearly combined multiple
  classifiers.
\newblock \emph{Pattern Recognition}, 37\penalty0 (6):\penalty0 1245--1265,
  2004.

\bibitem[Osband et~al.(2016)Osband, Blundell, Pritzel, and
  Van~Roy]{osband2016deep}
Ian Osband, Charles Blundell, Alexander Pritzel, and Benjamin Van~Roy.
\newblock Deep exploration via bootstrapped dqn.
\newblock In \emph{Advances in neural information processing systems}, pages
  4026--4034, 2016.

\bibitem[Lakshminarayanan et~al.(2017)Lakshminarayanan, Pritzel, and
  Blundell]{lakshminarayanan2017simple}
Balaji Lakshminarayanan, Alexander Pritzel, and Charles Blundell.
\newblock Simple and scalable predictive uncertainty estimation using deep
  ensembles.
\newblock In \emph{Advances in Neural Information Processing Systems}, pages
  6402--6413, 2017.

\bibitem[Snoek et~al.(2019)Snoek, Ovadia, Fertig, Lakshminarayanan, Nowozin,
  Sculley, Dillon, Ren, and Nado]{snoek2019can}
Jasper Snoek, Yaniv Ovadia, Emily Fertig, Balaji Lakshminarayanan, Sebastian
  Nowozin, D~Sculley, Joshua Dillon, Jie Ren, and Zachary Nado.
\newblock Can you trust your model's uncertainty? evaluating predictive
  uncertainty under dataset shift.
\newblock In \emph{Advances in Neural Information Processing Systems}, pages
  13969--13980, 2019.

\bibitem[Bishop(1994)]{bishop1994novelty}
Christopher~M Bishop.
\newblock Novelty detection and neural network validation.
\newblock \emph{IEE Proceedings-Vision, Image and Signal processing},
  141\penalty0 (4):\penalty0 217--222, 1994.

\bibitem[Markou and Singh(2003)]{markou2003novelty}
Markos Markou and Sameer Singh.
\newblock Novelty detection: a review—part 1: statistical approaches.
\newblock \emph{Signal processing}, 83\penalty0 (12):\penalty0 2481--2497,
  2003.

\bibitem[Senge et~al.(2014)Senge, B{\"o}sner, Dembczy{\'n}ski, Haasenritter,
  Hirsch, Donner-Banzhoff, and H{\"u}llermeier]{senge2014reliable}
Robin Senge, Stefan B{\"o}sner, Krzysztof Dembczy{\'n}ski, J{\"o}rg
  Haasenritter, Oliver Hirsch, Norbert Donner-Banzhoff, and Eyke
  H{\"u}llermeier.
\newblock Reliable classification: Learning classifiers that distinguish
  aleatoric and epistemic uncertainty.
\newblock \emph{Information Sciences}, 255:\penalty0 16--29, 2014.

\bibitem[Kukar and Kononenko(2002)]{kukar2002reliable}
Matja{\v{z}} Kukar and Igor Kononenko.
\newblock Reliable classifications with machine learning.
\newblock In \emph{European Conference on Machine Learning}, pages 219--231.
  Springer, 2002.

\bibitem[Geifman and El-Yaniv(2019)]{geifman2019selectivenet}
Yonatan Geifman and Ran El-Yaniv.
\newblock Selectivenet: A deep neural network with an integrated reject option.
\newblock \emph{arXiv preprint arXiv:1901.09192}, 2019.

\bibitem[Myers et~al.(2020)Myers, Ng, Severson, Kartoun, Dai, Huang, Anderson,
  and Stultz]{myers2020identifying}
Paul~D Myers, Kenney Ng, Kristen Severson, Uri Kartoun, Wangzhi Dai, Wei Huang,
  Frederick~A Anderson, and Collin~M Stultz.
\newblock Identifying unreliable predictions in clinical risk models.
\newblock \emph{NPJ Digital Medicine}, 3, 2020.

\bibitem[Jiang et~al.(2018)Jiang, Kim, Guan, and Gupta]{jiang2018trust}
Heinrich Jiang, Been Kim, Melody Guan, and Maya Gupta.
\newblock To trust or not to trust a classifier.
\newblock In \emph{Advances in neural information processing systems}, pages
  5541--5552, 2018.

\bibitem[Filos et~al.(2019{\natexlab{b}})Filos, Farquhar, Gomez, Rudner,
  Kenton, Smith, Alizadeh, de~Kroon, and Gal]{filos2019systematic}
Angelos Filos, Sebastian Farquhar, Aidan~N Gomez, Tim~GJ Rudner, Zachary
  Kenton, Lewis Smith, Milad Alizadeh, Arnoud de~Kroon, and Yarin Gal.
\newblock A systematic comparison of bayesian deep learning robustness in
  diabetic retinopathy tasks.
\newblock \emph{arXiv preprint arXiv:1912.10481}, 2019{\natexlab{b}}.

\bibitem[Kingma and Welling(2014)]{vae}
Diederik~P. Kingma and Max Welling.
\newblock Auto-encoding variational bayes.
\newblock In \emph{2nd International Conference on Learning Representations,
  {ICLR} 2014, Banff, AB, Canada, April 14-16, 2014, Conference Track
  Proceedings}, 2014.
\newblock URL \url{http://arxiv.org/abs/1312.6114}.

\bibitem[Efron(1979)]{bootstrap}
B.~Efron.
\newblock Bootstrap methods: Another look at the jackknife.
\newblock \emph{The Annals of Statistics}, 7\penalty0 (1):\penalty0 1--26,
  1979.
\newblock ISSN 00905364.
\newblock URL \url{http://www.jstor.org/stable/2958830}.

\bibitem[Naeini et~al.(2015)Naeini, Cooper, and Hauskrecht]{binningECE}
Mahdi~Pakdaman Naeini, Gregory Cooper, and Milos Hauskrecht.
\newblock Obtaining well calibrated probabilities using bayesian binning.
\newblock In \emph{Twenty-Ninth AAAI Conference on Artificial Intelligence},
  2015.

\bibitem[Hendrycks and Gimpel(2017)]{hendrycks17baseline}
Dan Hendrycks and Kevin Gimpel.
\newblock A baseline for detecting misclassified and out-of-distribution
  examples in neural networks.
\newblock \emph{Proceedings of International Conference on Learning
  Representations}, 2017.

\bibitem[Johnson et~al.(2016)Johnson, Pollard, Shen, Li-wei, Feng, Ghassemi,
  Moody, Szolovits, Celi, and Mark]{MIMIC}
Alistair~EW Johnson, Tom~J Pollard, Lu~Shen, H~Lehman Li-wei, Mengling Feng,
  Mohammad Ghassemi, Benjamin Moody, Peter Szolovits, Leo~Anthony Celi, and
  Roger~G Mark.
\newblock Mimic-iii, a freely accessible critical care database.
\newblock \emph{Scientific data}, 3:\penalty0 160035, 2016.

\bibitem[Purushotham et~al.(2017)Purushotham, Meng, Che, and
  Liu]{mimicbenchmark2}
Sanjay Purushotham, Chuizheng Meng, Zhengping Che, and Yan Liu.
\newblock Benchmark of deep learning models on large healthcare mimic datasets.
\newblock \emph{Journal of Biomedical Informatics}, 83, 10 2017.
\newblock \doi{10.1016/j.jbi.2018.04.007}.

\bibitem[Harutyunyan et~al.(2019)Harutyunyan, Khachatrian, Kale, Ver~Steeg, and
  Galstyan]{benchmarking}
Hrayr Harutyunyan, Hrant Khachatrian, David~C Kale, Greg Ver~Steeg, and Aram
  Galstyan.
\newblock Multitask learning and benchmarking with clinical time series data.
\newblock \emph{Scientific data}, 6\penalty0 (1):\penalty0 1--18, 2019.

\bibitem[Platt et~al.(1999)]{platt1999probabilistic}
John Platt et~al.
\newblock Probabilistic outputs for support vector machines and comparisons to
  regularized likelihood methods.
\newblock \emph{Advances in large margin classifiers}, 10\penalty0
  (3):\penalty0 61--74, 1999.

\bibitem[Paszke et~al.(2017)Paszke, Gross, Chintala, Chanan, Yang, DeVito, Lin,
  Desmaison, Antiga, and Lerer]{pytorch}
Adam Paszke, Sam Gross, Soumith Chintala, Gregory Chanan, Edward Yang, Zachary
  DeVito, Zeming Lin, Alban Desmaison, Luca Antiga, and Adam Lerer.
\newblock Automatic differentiation in pytorch.
\newblock 2017.

\bibitem[Pedregosa et~al.(2011)Pedregosa, Varoquaux, Gramfort, Michel, Thirion,
  Grisel, Blondel, Prettenhofer, Weiss, Dubourg, Vanderplas, Passos,
  Cournapeau, Brucher, Perrot, and Duchesnay]{scikit-learn}
F.~Pedregosa, G.~Varoquaux, A.~Gramfort, V.~Michel, B.~Thirion, O.~Grisel,
  M.~Blondel, P.~Prettenhofer, R.~Weiss, V.~Dubourg, J.~Vanderplas, A.~Passos,
  D.~Cournapeau, M.~Brucher, M.~Perrot, and E.~Duchesnay.
\newblock Scikit-learn: Machine learning in {P}ython.
\newblock \emph{Journal of Machine Learning Research}, 12:\penalty0 2825--2830,
  2011.

\end{thebibliography}

\newpage
\appendix
\section{OOD descriptives}
\label{app:OOD}

\begin{table}[ht]
\small
\caption{Characteristics and origin of the selected subgroups.}
\label{tab:ood}
\begin{tabular}{@{}lllll@{}}
\toprule
                                    & \textbf{Count} & \textbf{\begin{tabular}[c]{@{}l@{}}Mortality \\ rate (\%)\end{tabular}} & \textbf{MIMIC table} & \textbf{Field}  \\ \midrule
Emergency/Urgent admissions         & 18299          & 14.8                                                                    & ADMISSIONS           & ADMISSION\_TYPE \\
Elective admissions                 & 2840           & 3.4                                                                     & ADMISSIONS           & ADMISSION\_TYPE \\
Ethnicity: Asian                    & 492            & 13.8                                                                    & ADMISSIONS           & ETHNICITY       \\
Ethnicity: Black/African American   & 2016           & 9.2                                                                     & ADMISSIONS           & ETHNICITY       \\
Ethnicity: Hispanic/Latino          & 679            & 8.1                                                                     & ADMISSIONS           & ETHNICITY       \\
Ethnicity: White                    & 15043          & 12.9                                                                    & ADMISSIONS           & ETHNICITY       \\
Female                              & 9510           & 13.5                                                                    & ADMISSIONS           & GENDER          \\
Male                                & 11629          & 13                                                                      & ADMISSIONS           & GENDER          \\
Thyroid disorders                   & 1729           & 13.5                                                                    & DIAGNOSES\_ICD       & ICD9\_CODE      \\
Acute and unspecified renal failure & 4275           & 21.7                                                                    & DIAGNOSES\_ICD       & ICD9\_CODE      \\
Pancreatic disorders (not diabetes) & 319            & 14.1                                                                    & DIAGNOSES\_ICD       & ICD9\_CODE      \\
Epilepsy, convulsions               & 965            & 15.3                                                                    & DIAGNOSES\_ICD       & ICD9\_CODE      \\
Hypertension with complications   & 2274           & 15.1                                                                    & DIAGNOSES\_ICD       & ICD9\_CODE      \\ \bottomrule
\end{tabular}

\end{table}
\section{Model specifications}
\label{app:models}
\paragraph{NN ensemble} For the experiments on the MIMIC data we create an ensemble of 5 neural networks with two hidden layers of 100 neurons, ReLU activations, a dropout rate of $0.5$ and a sigmoid activation in the final layer. The NNs were all independently trained with the Adam optimizer, using default learning rate $1e^{-3}$, cross-entropy loss and a batch size of 256. We implemented early stopping and quit training when the validation loss did not improve for 2 epochs.
For the experiments on toy data we employed smaller networks with one hidden layer of 5 neurons, training with a smaller batch size of 8 and for 20 epochs. We implemented class weighting based on the relative class frequencies in each mini-batch. We calculated the positive weight for each batch $b$ as $w_b^+ = \frac{N^{-}_{b}}{N^{+}_{b}}$, that is, the number of negative examples divided by the number of positive examples in the mini-batch. This results in the following weighed loss $\mathcal{L}_b$ for the mini-batch:

\begin{equation}
    \mathcal{L}_{b} = - \frac{1}{N_{b}} \sum_{i=1}^{N_b} \left[ w_b^+ \cdot y_{i} \log \hat{y_i}_{} + (1-y_{i})\log(1-\hat{y}_i)\right]
\end{equation}
where $N_b$ is the number of data points in the batch, $y_i$ the true label and $\hat{y}_i$ the predicted probability. 
We implemented the NN ensemble in PyTorch \citep{pytorch}.

\paragraph{MC Dropout} For all experiments we use the same architecture and training procedure as described for the NN ensemble, but train only a single network. During inference time we take the average over 100 forward passes with the Dropout layers activated.

\paragraph{Bootstrapped Logistic Regression}
We train an ensemble of 5 models, training each model on a different bootstrapped sample of the data of the same size as the training data. We used the scikit-learn implementation of Logistic Regression  \citep{scikit-learn} with the L-BFGS-B solver. We set the regularization coefficient to $C=1e^{-2}$ on real data and to $0$ on toy data.
For class weighting we used the same weighting as described for the Deep Ensemble, but we employed a constant positive weight based on all data instead of one that is batch-dependent. 
\paragraph{Variational Autoencoder}
We implemented the VAE in PyTorch \citep{pytorch}. The latent distribution was modeled as a 500-dimensional normal distribution with diagonal covariance. The output distribution of the decoder was also modeled as a normal distribution with diagonal covariance.  We trained the model with Adam optimizer, again using a default learning rate of $1e^{-3}$. The batch size was set to 256 and training lasted 30 epochs. The encoder and decoder both had no hidden layer, as adding hidden layers made training very unstable and did not seem to improve performance significantly.
During test time, we took 10 samples from the latent distribution and calculated the average reconstruction error over these sample. This reconstruction error is what we use as a measure of uncertainty. 
\section{Toy data generation}
\label{app:toy}
For the balanced data examples, we generated two equally-sized clusters, one for each label, both from a multivariate normal distribution. The means of the positive and negative clusters were $(2, 2)$ and $(-1, -1)$ respectively. Both clusters had diagonal covariance matrices defined as 4 times the identity matrix.  For the unbalanced data set, we used the same multivariate normal distribution for the negative cluster, and for the positive cluster the same mean $(2,2)$ with a covariance of 2 times the identity matrix. We generated 6 times as many points in the negative cluster as compared to the positive cluster. In both cases we generated 200 points for training. 

\section{Extra plots on toy data}
\label{app:extratoy}
\begin{figure}[h]
  \subfloat[]{
	\begin{minipage}[c][1\width]{
	   0.32\textwidth}
	   \centering
	   \includegraphics[width=1\linewidth]{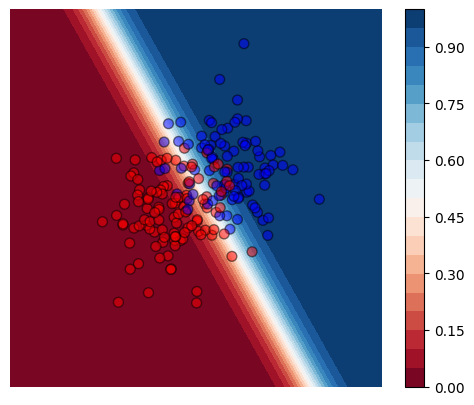}
	   \label{fig:lr_single}
	\end{minipage}}
 \hfill 	
  \subfloat[]{
	\begin{minipage}[c][1\width]{
	   0.32\textwidth}
	   \centering
	   \includegraphics[width=1\linewidth]{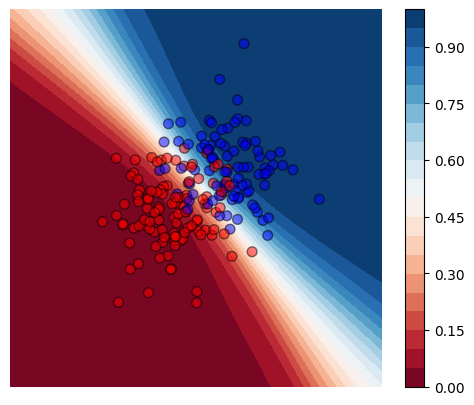}
	   \label{fig:lr_ensemble}
	\end{minipage}}
	\hrulefill
	  \subfloat[]{
	\begin{minipage}[c][1\width]{
	   0.32\textwidth}
	   \centering
	   \includegraphics[width=1\linewidth]{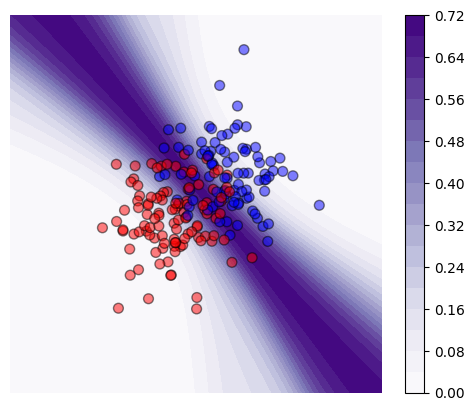}
	   \label{fig:lr_ensemble_entr}
	\end{minipage}}
\caption{Showing the predicted probabilities of a single LR model (a), the average of the predicted probabilities of bootstrapped LR models (b) and the entropy of these predicted probabilities (c).}
\label{fig:lr_example}
\end{figure}

\begin{figure}[h]
  \subfloat[]{
	\begin{minipage}[c][1\width]{
	   0.32\textwidth}
	   \centering
	   \includegraphics[width=1\linewidth]{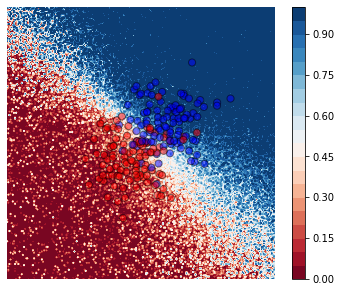}
	   \label{fig:mc_single}
	\end{minipage}}
 \hfill 	
  \subfloat[]{
	\begin{minipage}[c][1\width]{
	   0.32\textwidth}
	   \centering
	   \includegraphics[width=1\linewidth]{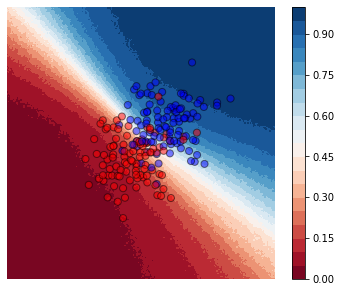}
	   \label{fig:mc_ensemble}
	\end{minipage}}
	\hrulefill
	  \subfloat[]{
	\begin{minipage}[c][1\width]{
	   0.32\textwidth}
	   \centering
	   \includegraphics[width=1\linewidth]{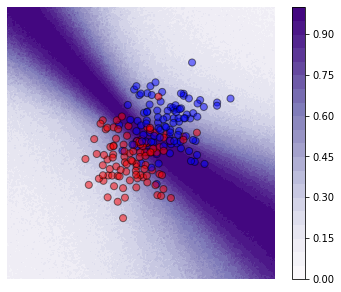}
	   \label{fig:mc_ensemble_entr}
	\end{minipage}}
\caption{Showing the predicted probabilities of a single forward pass with dropout (a), the average of 100 forward passes (b) and the entropy of these predicted probabilities (c).}
\label{fig:mc_example}
\end{figure}


\end{document}